\documentclass[nopreprintline,12pt]{templates/elsarticle}

\usepackage{amssymb}

\usepackage{soul,color}
\usepackage{url}
\usepackage{amsmath}
\usepackage{siunitx}
\usepackage{textcomp}
\usepackage{makecell}
\usepackage{multirow}
\usepackage{wrapfig}
\usepackage[noend]{algorithm2e}
\usepackage{hyperref}
\usepackage{fancyhdr}

\journal{Neurocomputing}

\begin{document}

\begin{frontmatter}


\title{Pseudo-Rehearsal: Achieving Deep Reinforcement Learning without Catastrophic Forgetting}
\author{Craig~Atkinson\corref{cor1}\fnref{label2}}
\ead{atkcr398@student.otago.ac.nz}
\author{Brendan~McCane\fnref{label2}}
\author{Lech~Szymanski\fnref{label2}}
\author{Anthony~Robins\fnref{label2}}
\fntext[label2]{Department of Computer Science, University of Otago, 133 Union Street East, Dunedin, New Zealand\newline \newline Accepted for publication in Neurocomputing (\url{https://doi.org/10.1016/j.neucom.2020.11.050}). $\copyright$ 2020. This manuscript version is made available under the CC-BY-NC-ND 4.0 license \url{http://creativecommons.org/licenses/by-nc-nd/4.0/}.}
\cortext[cor1]{Corresponding author.}



\author{}

\address{}

\begin{abstract}
Neural networks can achieve excellent results in a wide variety of applications. However, when they attempt to sequentially learn, they tend to learn the new task while catastrophically forgetting previous ones. We propose a model that overcomes catastrophic forgetting in sequential reinforcement learning by combining ideas from continual learning in both the image classification domain and the reinforcement learning domain. This model features a dual memory system which separates continual learning from reinforcement learning and a pseudo-rehearsal system that ``recalls'' items representative of previous tasks via a deep generative network. Our model sequentially learns Atari $2600$ games without demonstrating catastrophic forgetting and continues to perform above human level on all three games. This result is achieved without: demanding additional storage requirements as the number of tasks increases, storing raw data or revisiting past tasks. In comparison, previous state-of-the-art solutions are substantially more vulnerable to forgetting on these complex deep reinforcement learning tasks.
\end{abstract}




\end{frontmatter}


\section{Introduction}
There has been enormous growth in research around reinforcement learning since the development of Deep Q-Networks (DQNs)~\cite{mnih2015human}. DQNs apply Q-learning to deep networks so that complicated reinforcement tasks can be learnt. However, as with most distributed models, DQNs can suffer from Catastrophic Forgetting (CF)~\cite{mccloskey1989catastrophic, kirkpatrick2017overcoming}. This is where a model has the tendency to forget previous knowledge as it learns new knowledge. Pseudo-rehearsal is a method for overcoming CF by rehearsing randomly generated examples of previous tasks, while learning on real data from a new task. Although pseudo-rehearsal methods have been widely used in image classification, they have been virtually unexplored in reinforcement learning. Solving CF in the reinforcement learning domain is essential if we want to achieve artificial agents that can continuously learn.

Continual learning is important to neural networks because CF limits their potential in numerous ways. For example, imagine a previously trained network whose function needs to be extended or partially changed. The typical solution would be to train the neural network on all of the previously learnt data (that was still relevant) along with the data to learn the new function. This can be an expensive operation because previous datasets (which tend to be very large in deep learning) would need to be stored and retrained. However, if a neural network could adequately perform continual learning, it would only be necessary for it to directly learn on data representing the new function. Furthermore, continual learning is also desirable because it allows the solution for multiple tasks to be compressed into a single network where weights common to both tasks may be shared. This can also benefit the speed at which new tasks are learnt because useful features may already be present in the network.

Our Reinforcement-Pseudo-Rehearsal model (RePR\footnote{Read as ``reaper".})  achieves continual learning in the reinforcement domain. It does so by utilising a dual memory system where a freshly initialised DQN is trained on the new task and then knowledge from this short-term network is transferred to a separate DQN containing long-term knowledge of all previously learnt tasks. A generative network is used to produce states (short sequences of data) representative of previous tasks which can be rehearsed while transferring knowledge of the new task. For each new task, the generative network is trained on pseudo-items produced by the previous generative network, alongside data from the new task. Therefore, the system can prevent CF without the need for a large memory store holding data from previously encountered training examples.

The reinforcement tasks learnt by RePR are Atari $2600$ games. These games are considered complex because the input space of the games is large which currently requires reinforcement learning to use deep neural networks (i.e. deep reinforcement learning). Applying pseudo-rehearsal methods to deep reinforcement learning is challenging because these reinforcement learning methods are notoriously unstable compared to image classification (due to the deadly triad~\cite{sutton2017reinforcement}). In part, this is because target values are consistently changing during learning. We have found that using pseudo-rehearsal while learning these non-stationary targets is difficult because it increases the interference between new and old tasks. Furthermore, generative models struggle to produce high quality data resembling these reinforcement learning tasks, which can prevent important task knowledge from being learnt for the first time, as well as relearnt once it is forgotten.



Our RePR model applies pseudo-rehearsal to the difficult domain of deep reinforcement learning. RePR introduces a dual memory model suitable for reinforcement learning. This model is novel compared to previously used dual memory pseudo-rehearsal models in two important aspects. Firstly, the model isolates reinforcement learning to the short-term system, so that the long-term system can use supervised learning (i.e. mean squared error) with fixed target values (converged on by the short-term network), preventing non-stationary target values from increasing the interference between new and old tasks. Importantly, this differs from previous applications of pseudo-rehearsal, where both the short-term and long-term systems learn with the same cross-entropy loss function. Secondly, RePR transfers knowledge between the dual memory system using real samples, rather than those produced by a generative model. This allows tasks to be learnt and retained to a higher performance in reinforcement learning. The source code for RePR can be found at \url{https://bitbucket.org/catk1ns0n/repr_public/}.

A summary of the main contributions of this paper are:

\begin{itemize}
  \item the first successful application of pseudo-rehearsal methods to complex deep reinforcement learning tasks;
  \item above state-of-the-art performance when sequentially learning complex reinforcement tasks, without storing any raw data from previously learnt tasks;
  \item empirical evidence demonstrating the need for a dual memory system as it facilitates new learning by separating the reinforcement learning system from the continual learning system.
\end{itemize}

\section{Background}

\subsection{Deep Q-Learning}

In deep Q-learning~\cite{mnih2015human}, the neural network is taught to predict the discounted reward that would be received from taking each one of the possible actions given the current state.  More specifically, it minimises the following loss function:
\begin{equation}
\label{dqn-loss}
L_{DQN} = \mathbb{E}_{(s_t,a_t,r_t,d_t,s_{t+1})\sim U(D)}\Bigg[\bigg(y_t-Q(s_t,a_t;\psi_t)\bigg)^2\Bigg],
\end{equation} \begin{equation}
    y_t= 
\begin{cases}
    r_t,& \text{if terminal at } t + 1\\
    r_t + {\gamma}\max\limits_{a_{t+1}}Q(s_{t+1},a_{t+1};\psi_t^-),              & \text{otherwise}
\end{cases}
\end{equation} where there exist two $Q$ functions, a deep predictor network and a deep target network. The predictor's parameters $\psi_t$ are updated continuously by stochastic gradient descent and the target's parameters $\psi_t^-$ are infrequently updated with the values of $\psi_t$. The tuple $(s_t,a_t,r_t,d_t,s_{t+1})\sim U(D)$ consists of the state, action, reward, terminal and next state for a given time step $t$ drawn uniformly from a large record of previous experiences, known as an experience replay.

\subsection{Pseudo-Rehearsal}

The simplest way of solving the CF problem is to use a rehearsal strategy, where previously learnt items are practised alongside the learning of new items. Researchers have proposed extensions to rehearsal~\cite{lopez2017gradient,rebuffi2017icarl,riemer2018generative}.  However, all these rehearsal methods still have disadvantages, such as requiring excessive amounts of data to be stored from previously seen tasks\footnote{An experience replay differs from rehearsal because it only stores recently seen data from the current task and therefore, learning data from the experience replay does not prevent forgetting of previously seen tasks.}. Furthermore, in certain applications (e.g. the medical field), storing real data for rehearsal might not be possible due to privacy regulations. Additionally, rehearsal is not biologically plausible - mammalian brains don't retain raw sensory information over the course of their lives.  Hence, cognitive research might be a good inspiration for tackling the CF problem.


Pseudo-rehearsal was proposed as a solution to CF which does not require storing large amounts of past data and thus, overcomes the previously mentioned shortcomings of rehearsal~\cite{robins1995catastrophic}. Originally, pseudo-rehearsal involved constructing a pseudo-dataset by generating random inputs (i.e. from a random number generator), passing them through the original network and recording their output. This meant that when a new dataset was learnt, the pseudo-dataset could be rehearsed alongside it, resulting in the network learning the data with minimal changes to the previously modelled function.

There is psychological research that suggests that mammal brains use an analogous method to pseudo-rehearsal to prevent CF in memory consolidation. Memory consolidation is the process of transferring memory from the hippocampus, which is responsible for short-term knowledge, to the cortex for long-term storage. The hippocampus and sleep have both been linked as important components for retaining previously learnt information~\cite{gais2007sleep}. The hippocampus has been observed to replay patterns of activation that occurred during the day while sleeping~\cite{louie2001temporally}, similar to the way that pseudo-rehearsal generates previous experiences. Therefore, we believe that pseudo-rehearsal based mechanisms are neurologically plausible and could be useful as an approach to solving the CF problem in deep reinforcement learning.

Plain pseudo-rehearsal does not scale well to data with large input spaces such as images~\cite{atkinson2018pseudo}. This is because the probability of a population of randomly generated inputs sampling the space of possible input images with sufficient density is essentially zero. This is where Deep Generative Replay~\cite{shin2017continual} and Pseudo-Recursal~\cite{atkinson2018pseudo} have leveraged Generative Adversarial Networks (GANs)~\cite{goodfellow2014generative} to randomly generate pseudo-items representative of previously learnt items.

A GAN has two components; a generator and a discriminator. The discriminator is trained to distinguish between real and generated images, whereas the generator is trained to produce images which fool the discriminator. When a GAN is used alongside pseudo-rehearsal, the GAN is also trained on the task so that its generator learns to produce items representative of the task's input items. Then, when a second task needs to be learnt, pseudo-items can be generated randomly from the GAN's generator and used in pseudo-rehearsal. More specifically, the minimised cost for pseudo-rehearsal is:
\begin{equation}
J = \mathcal{L}(h\left(x_i; \theta_i), y_i\right) + \sum\limits_{j = 1}^{i-1} \mathcal{L}\left(h(\widetilde{x}_j; \theta_i), \widetilde{y}_j\right),
\end{equation} where $\mathcal{L}$ is a loss function,  such as cross-entropy, and $h$ is a neural network with weights $\theta_i$ while learning task $i$. The tuple $x_i, y_i$ is the input-output pair for the current task. The tuple  $\widetilde{x}_j, \widetilde{y}_{j}$ is the input-output pair for a pseudo-item, where $\widetilde{x}_j$ is the input generated so that it represents the previous task $j$ and $\widetilde{y}_{j}$ is the target output calculated by $\widetilde{y}_{j} = h(\widetilde{x}_j; \theta_{i-1})$.

This technique can be applied to multiple tasks using only a single GAN by doing pseudo-rehearsal on the GAN as well. Thus, the GAN learns to generate items representative of the new task while still remembering to generate items representative of the previous tasks (by rehearsing the pseudo-items it generates). This technique has been shown to be very effective for remembering a chain of multiple image classification tasks~\cite{shin2017continual,atkinson2018pseudo}.

\section{The RePR Model}

RePR is a dual memory model which uses pseudo-rehearsal with a generative network to achieve sequential learning in reinforcement tasks. The first part of our dual memory model is the short-term memory (STM) system\footnote{Note that the interpretation of the term ``short-term memory" (STM) as used here and in related literature varies somewhat from the use of the same term in the general psychological literature.}, which serves a role analogous to that of the hippocampus in learning and is used to learn the current task. The STM system contains two components; a DQN that learns the current task and an experience replay containing data only from the current task. The second part is the long-term memory (LTM) system, which serves a role analogous to that of the cortex. The LTM system also has two components; a DQN containing knowledge of all tasks learnt and a GAN which can generate states representative of these tasks. During consolidation, the LTM system retains previous knowledge through pseudo-rehearsal, while being taught by the STM system how to respond on the current task.

Transferring knowledge between these two systems is achieved through knowledge distillation~\cite{hinton2015distilling}, where a student network is optimised so that it outputs similar values to a teacher network. In the RePR model, the student network is the long-term DQN and the teacher network is the short-term DQN. The key difference between distillation and pseudo-rehearsal is that distillation uses real items to teach new knowledge, whereas pseudo-rehearsal uses generated items to retain previously learnt knowledge.

In reinforcement learning (deep Q-learning), the target Q-values are from a non-stationary distribution because they are produced by the target network which is being updated during training. Our dual memory system in RePR is important because it allows the difficult job of reinforcement learning to be isolated to the short-term DQN, so that no other learning can interfere with it. Consequently, this reduces the difficulty of the problem for the long-term DQN as it does not need to learn this non-stationary distribution while also trying to retain knowledge of previous tasks. Instead, the long-term DQN can use supervised learning to learn the final stationary distribution Q-values, while rehearsing previous knowledge through pseudo-rehearsal.

\subsection{Training Procedure}

The training procedure can be broken down into three steps: short-term DQN training, long-term DQN training and long-term GAN training. This process could be repeated for any number of tasks until the DQN or GAN run out of capacity to perform the role sufficiently.

\subsubsection{Training the Short-Term DQN}

\begin{figure}[t]
\vskip 0.2in
\begin{center}
\centerline{\includegraphics[trim={0.4cm 1cm 11cm 0.5cm}, clip, width=\columnwidth]{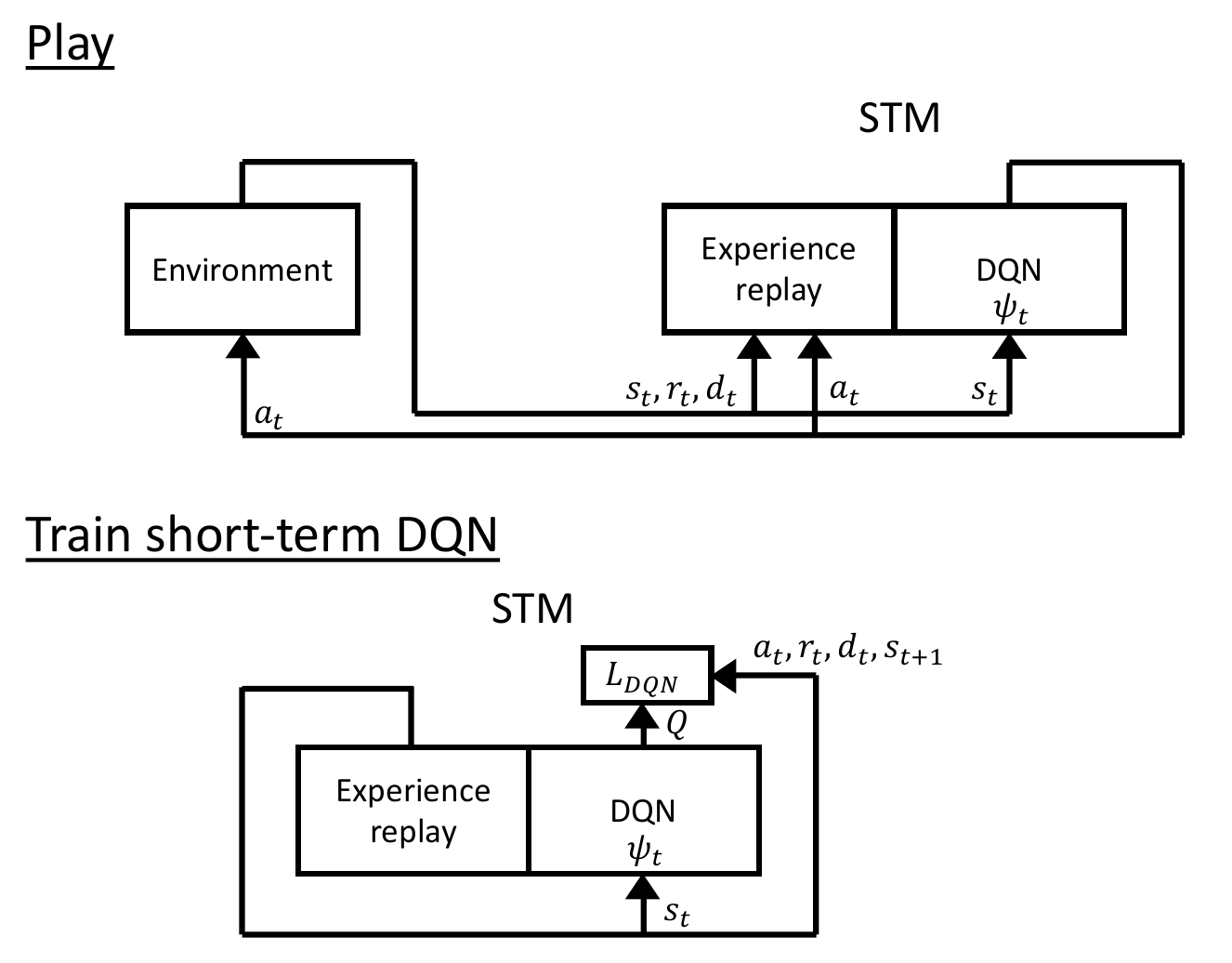}}
\caption{Flow of information while training the STM system. The model plays the game while simultaneously training the STM system.}
\label{STM-training}
\end{center}
\vskip -0.2in
\end{figure}

When there is a new task to be learnt, the short-term DQN is reinitialised and trained solely on the task using the standard DQN loss function (Equation~\ref{dqn-loss}). A summary of the information flow while training the STM system can be found in Fig.~\ref{STM-training} and the related pseudo-code can be found in the appendices as Algorithm~\ref{stm-alg}.

\subsubsection{Training the Long-Term DQN}

\begin{figure}[t]
\vskip 0.2in
\begin{center}
\centerline{\includegraphics[trim={0cm 0cm 0cm 0cm}, clip, width=\columnwidth]{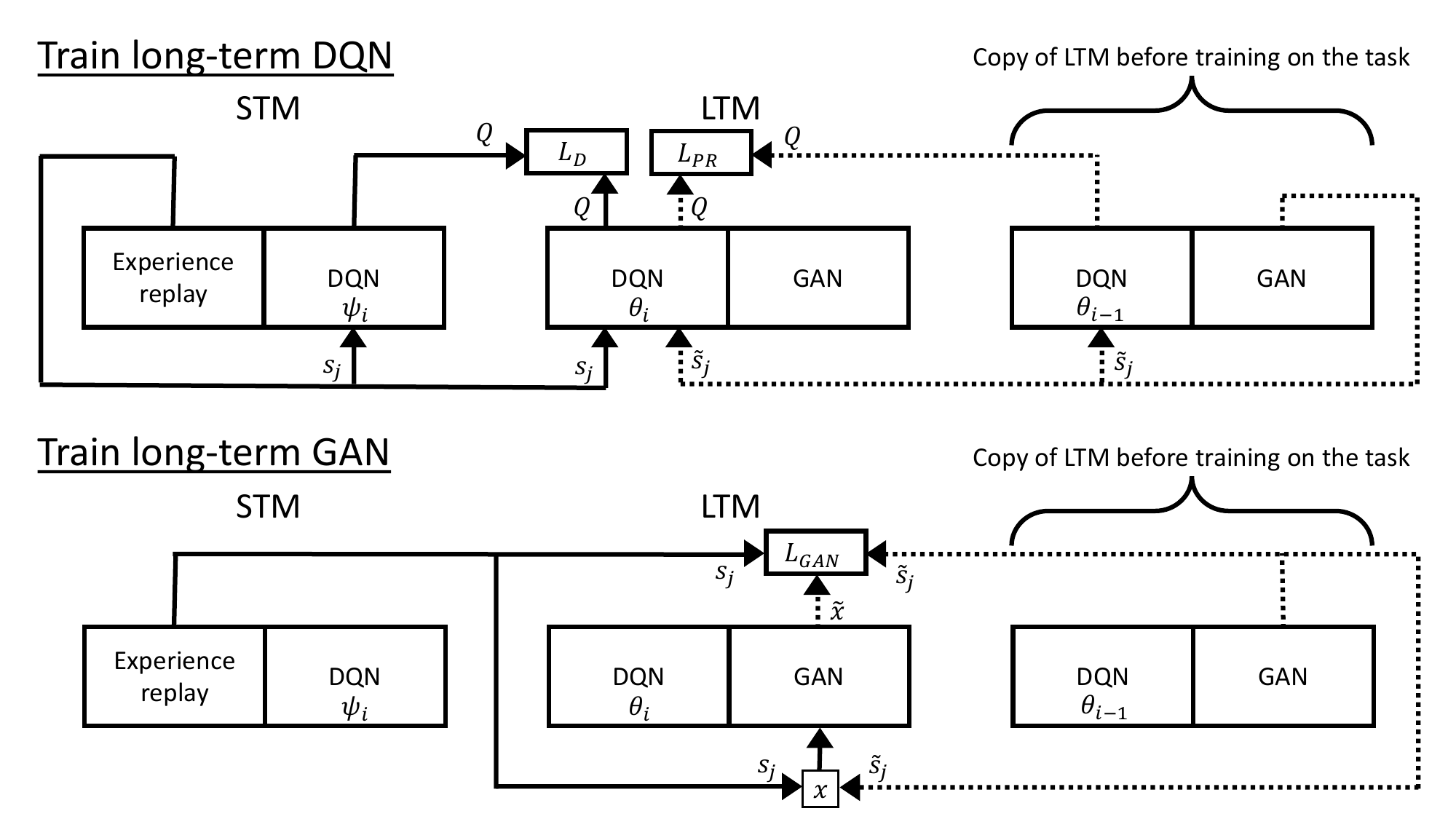}}
\caption{Flow of information while training the LTM system. Solid lines represent the information flow for learning the new task and dashed lines represent the information flow for retaining previous tasks with pseudo-rehearsal. The DQN and GAN are both trained at independent times.}
\label{LTM-training}
\end{center}
\vskip -0.2in
\end{figure}

Knowledge is transferred to the long-term DQN by teaching it to produce similar outputs to the short-term DQN on examples from its experience replay. Alongside this, the long-term DQN is constrained with pseudo-rehearsal to produce similar output values to the previous long-term DQN on states generated from the long-term GAN. More specifically, the loss functions used are:
\begin{equation}
\label{repr-loss}
L_{LTM} =  \frac 1 N \sum\limits_{j = 1}^{N}\alpha {L_D}_j + (1-\alpha) {L_{PR}}_j,
\end{equation}
\begin{equation}
\label{distill-loss}
{L_D}_j = \sum\limits_{a}^{A}(Q(s_j,a;\theta_i) - Q(s_j,a;\psi_i))^2,
\end{equation}
\begin{equation}
\label{pr-loss}
{L_{PR}}_j =\sum\limits_{a}^{A}(Q(\widetilde{s}_j,a;\theta_i) - Q(\widetilde{s}_j,a;\theta_{i-1}))^2,
\end{equation} where $s_j$ is a state drawn from the current task's experience replay, $N$ is the mini-batch size, $A$ is the set of possible actions, $\theta_i$ is the long-term DQN's weights on the current task, $\psi_i$ is the short-term DQN's weights after learning the current task and $\theta_{i-1}$ is the long-term DQN's weights after learning the previous task. Pseudo-items' inputs $\widetilde{s}_j$ are generated from a GAN and are representative of states in previously learnt games. The trade off between learning the new tasks and retaining the previous ones is controlled by a scaling factor ($0 < \alpha < 1$). A summary of the information flow while training the long-term DQN can be found in Fig.~\ref{LTM-training} and the related pseudo-code can be found in the appendices as Algorithm~\ref{ltm-alg}.

\subsubsection{Training the Long-Term GAN}

The GAN is reinitialised and trained to produce both states that are representative of states the previous GAN outputs and states drawn from the current task's experience replay. More specifically, the states the GAN learns are drawn such that: 
\begin{equation}
\label{gan-data}
    x= 
\begin{cases}
    s_j,& \text{if } r < \frac 1 T\\
    \widetilde{s}_j,              & \text{otherwise}
\end{cases}
\end{equation} where $r$ is a random number uniformly drawn from $[0, 1)$ and $s_j$ is a randomly selected state in the current task's experience replay. $T$ is the number of tasks learnt and $\widetilde{s}_j$ is a randomly generated state from the long-term GAN before training on task $i$. This state is a pseudo-item which ideally resembles a state from one of the prior tasks (task $1$ to $i-1$). The GAN in our experiments is trained with the WGAN-GP loss function~\cite{gulrajani2017improved} with a drift term~\cite{karras2017progressive}. The specific loss function can be found in~\ref{append-GAN}. A summary of the information flow while training the long-term GAN can be found in Fig.~\ref{LTM-training} and the related pseudo-code can be found in the appendices as Algorithm~\ref{gan-alg}.

\subsection{Requirements of a Continual Learning Agent}

We seek to operate within the constraints that apply to biological agents. As such, we seek a continual learning agent capable of learning multiple tasks sequentially: without revisiting them; without substantially forgetting previous tasks; with a consistent memory size that does not grow as the number of tasks increases; and without storing raw data from previous tasks. The results in Section~\ref{results} demonstrate that our RePR model can sequentially learn multiple tasks, without revisiting them or substantially forgetting previous tasks. Furthermore, our model does not store raw data because applying pseudo-rehearsal methods to a single GAN allows it to learn to generate data representative of all previous tasks, without growing in size as the number of tasks increases.


\section{Related Work}

This section will focus on methods for preventing CF in reinforcement learning and will generally concentrate on how to learn a new policy without forgetting those previously learnt for different tasks. There is a lot of related research outside of this domain (see~\cite{parisi2019continual} for a broad review), predominantly around continual learning in image classification. However, because these methods cannot be directly applied to complex reinforcement learning tasks, we have excluded them from this review.

There are two main strategies for avoiding CF. The first of these strategies is restricting how the network is optimised. Some models, such as Progressive neural networks~\cite{rusu2016progressive}, introduce a large number of units, or even separate networks, which are restrictively trained only on a particular task. Although these methods can share some weights, they still have a large number of task specific ones and thus, the model size needs to grow substantially as it continually learns.

Another way which the network's optimisation can be restricted is through weight constraints. These methods amend the loss function so that weights do not change considerably when learning a new task. The most popular of these methods is Elastic Weight Consolidation (EWC)~\cite{kirkpatrick2017overcoming}, which augments a network's loss function with a constraint that forces the network's weights to yield similar values to previous networks. Weights that are more important to the previous task/s are constrained more so that less important weights can be used to learn the new task. EWC has been paired with a DQN to learn numerous Atari $2600$ games. One undesirable requirement of EWC is that the network's weights after learning each task must be stored along with either the Fisher information matrix for each task or examples from past tasks (so that the matrix can be calculated when needed). Other variations for constraining the weights have also been proposed~\cite{kobayashi2018check,kaplanis2018continual}. However, these variations have only been applied to relatively simple reinforcement tasks.

Progress and Compress~\cite{schwarz2018progress} learnt multiple Atari games by firstly learning the game in STM and then using distillation to transfer it to LTM. The LTM system held all previously learnt tasks, counteracting CF using a modified version of EWC called online-EWC. This modified version does not scale in memory requirements as the number of tasks increases. This is because the algorithm stores only the weights of the network after learning the most recent task, along with the discounted sum of previous Fisher information matrices to use when constraining weights. In Progress and Compress, there are also layer-wise connections between the two systems to encourage the short-term network to learn using features already learnt by the long-term network.

The second main strategy for avoiding CF is to amend the training data to be more representative of previous tasks. This category includes both rehearsal and pseudo-rehearsal, as these methods add either real or generated samples to the training dataset. One example of a rehearsal method in reinforcement learning is PLAID~\cite{berseth2018progressive}. It uses distillation to merge a network that performs the new task with a network whose policy performs all previously learnt tasks. Distillation methods have also been applied to Atari $2600$ games~\cite{rusu2015policy, parisotto2015actor}. However, these were in multi-task learning where CF is not an issue. The major disadvantage with all rehearsal methods is that they require either having continuous access to previous environments or storing a large amount of previous training data. For other recent examples of rehearsal in reinforcement learning see~\cite{rolnick2018experience, isele2018selective, riemer2018learning}.

To our knowledge, pseudo-rehearsal has only been applied by~\cite{caselles2018continual} to sequentially learn reinforcement tasks. This was achieved by extending the Deep Generative Replay algorithm from image classification to reinforcement learning. Pseudo-rehearsal was combined with a Variational Auto-Encoder so that two very simple reinforcement tasks could be sequentially learnt by State Representation Learning without CF occurring. These tasks involved a 2D world where the agent's input was a small $64\times3$ grid representing the colour of objects it could see in front of it. The only thing that changed between tasks was the colour of the objects the agent must collect. This sequential learning environment is simple compared to the one we use and thus, there are a number of important differences in our RePR model. Our complex reinforcement learning tasks are relatively different from one another and have a large input space. This requires RePR to use deep convolutional networks, specifically DQNs and GANs, to learn and generate plausible input items. Furthermore, our model incorporates a dual memory system which isolates reinforcement learning to the STM system, improving the acquisition of the new task.

Without an experience replay, CF can occur while learning even a single task as the network forgets how to act in previously seen states. Pseudo-rehearsal has also been applied to this problem by rehearsing randomly generated input items from basic distributions (e.g. uniform distribution)~\cite{marochko2017pseudorehearsalvalue, baddeley2008reinforcement}, with a similar idea accomplished in actor-critic networks~\cite{marochko2017pseudorehearsalactor}. However, all these methods were applied to simple reinforcement tasks and did not utilise deep generative structures for producing pseudo-items or convolutional network architectures. Since our work, pseudo-rehearsal has been used to overcome CF in models which have learnt to generate states from previously seen environments~\cite{caselles2019, ketz2019using}. But in both these cases, pseudo-rehearsal was not applied to the learning agent to prevent CF.

The model which most closely resembles RePR is the Deep Generative Dual Memory Network (DGDMN)~\cite{kamra2017deep} used in sequential image classification. The DGDMN extends Deep Generative Replay by introducing a dual memory system similar to RePR. The STM system comprises one or more modules made up of a classifier paired with a generative network (i.e. a Variational Auto-Encoder). A separate module is used to learn each of the recent tasks. The LTM system comprises a separate classifier and generative network. When consolidation occurs, knowledge from the short-term modules is transferred to the LTM system while the long-term generator is used to produce pseudo-items for pseudo-rehearsal. 

The primary difference between DGDMN and RePR relates to the data used to teach the new task to the LTM system. The short-term modules in DGDMN each contain a generative network which learns to produce input examples representative of the new task. These examples are labelled by their respective classifier and then taught to the LTM system. However, in RePR, the STM system does not contain a generative network and instead uses data from its experience replay to train the LTM system. This experience replay does not grow in memory as it only contains a limited amount of data from the most recent task and as this data is more accurate than generated data, it is more effective in teaching the LTM system the new task. There are also key differences in the loss functions used for DGDMN and RePR. DGDMN was developed for sequential image classification and so, it uses the same cross-entropy loss function for learning new tasks in the STM system as in the LTM system. However, in RePR the learning process is separated. The more difficult reinforcement learning (deep Q-learning) is accomplished in the STM system. This is isolated from the LTM system, which can learn and retain through supervised learning (i.e. mean squared error).

To our knowledge, a dual memory system has not previously been used to separate different types of learning in pseudo-rehearsal algorithms. Although a similar dual memory system to RePR was used in Progress and Compress~\cite{schwarz2018progress}, unlike in RePR, authors of Progress and Compress did not find conclusive evidence that it assisted sequential reinforcement learning.

In real neuronal circuits it is a matter of debate whether memory is retained through synaptic stability, synaptic plasticity, or a mixture of mechanisms~\cite{abraham2005memory,gallistel2013neuroscience,abraham2019plasticity}. The synaptic stability hypothesis states that memory is retained through fixing the weights between units that encode it. The synaptic plasticity hypothesis states that the weights between the units can change as long as the output units still produce the correct output pattern. EWC and Progress and Compress are both methods that constrain the network's weights and therefore, align with the synaptic stability hypothesis. Pseudo-rehearsal methods amend the training dataset, pressuring the network's outputs to remain the same, without constraining the weights and therefore, align with the synaptic plasticity hypothesis. Pseudo-rehearsal methods have not yet been successfully applied to complex reinforcement learning tasks. The major advantage of methods that align with the synaptic plasticity hypothesis is that, when consolidating new knowledge, they allow the network to restructure its weights and compress previous representations to make room for new ones. This suggests that, in the long run, pseudo-rehearsal will outperform state-of-the-art weight constraint methods in reinforcement learning, since it allows neural networks to internally reconfigure in order to consolidate new knowledge.

In summary, RePR is the first variation of pseudo-rehearsal to be successfully applied to continual learning with complex reinforcement tasks.


\section{Method}

\begin{table*}[p]
\caption{Details of the experimental conditions.}
\label{exp-conds}
\renewcommand{\arraystretch}{1.5}
\begin{center}
\begin{scriptsize}
\makebox[\columnwidth]{
\begin{tabular}{c | c}
\hline
Condition & Description\\
\hline
$no\text{-}reh$ & \makecell[l]{For each new task, the task is learnt in the STM system using deep Q-learning (Equation~\ref{dqn-loss}).\\The task is then transferred to the LTM system using distillation (Equation~\ref{distill-loss}). This\\condition does not attempt to retain previously learnt tasks.}\\
\hline
$reh$ & \makecell[l]{For each new task, the task is learnt in the STM system using deep Q-learning (Equation~\ref{dqn-loss}).\\The task is then transferred to the LTM system using distillation, while previously learnt\\tasks are retained with rehearsal (Equation~\ref{repr-loss}, where the generated items $\widetilde{s}_j$ are replaced with\\real items from previous tasks).}\\
\hline
$RePR$ & \makecell[l]{For each new task, the task is learnt in the STM system using deep Q-learning (Equation~\ref{dqn-loss}).\\The task is then transferred to the long-term DQN using distillation, while previously learnt\\tasks are retained with pseudo-rehearsal (Equation~\ref{repr-loss}). The long-term GAN is then taught to\\generate data from the new task and previously learnt tasks (Equation~\ref{gan-data}) using the loss\\functions defined in Equation~\ref{disc-loss} and Equation~\ref{gen-loss} in~\ref{append-GAN}.}\\
\hline
$ewc$ & \makecell[l]{For each new task, the task is learnt in the STM system using deep Q-learning (Equation~\ref{dqn-loss}).\\The task is then transferred to the LTM system using distillation, while previously learnt\\tasks are retained with EWC (Equation~\ref{ltm-ewc-loss} in~\ref{append-EWC}).}\\
\hline
$online\text{-}ewc$ & \makecell[l]{For each new task, the task is learnt in the STM system using deep Q-learning (Equation~\ref{dqn-loss}).\\The task is then transferred to the LTM system using distillation (Equation~\ref{distill-loss}), while\\previously learnt tasks are retained with online-EWC (Equation~\ref{oewc-loss} in~\ref{append-EWC}).}\\
\hline
$PR$ & \makecell[l]{For each new task, the task is learnt in a DQN using deep Q-learning (Equation~\ref{dqn-loss}), while\\previously learnt tasks are retained with pseudo-rehearsal (Equation~\ref{pr-loss}, where the weights $\theta_i$\\are the same weights $\psi_t$ used in deep Q-learning). The GAN is then taught to generate data\\from the new task and previously learnt tasks (Equation~\ref{gan-data}) using the loss functions defined in\\Equation~\ref{disc-loss} and Equation~\ref{gen-loss} in~\ref{append-GAN}.}\\
\hline
$DGDMN$ & \makecell[l]{For each new task, the task is learnt in the short-term DQN using deep Q-learning\\(Equation~\ref{dqn-loss}). The short-term GAN is then taught to generate data from the new task using\\the loss functions defined in Equation~\ref{disc-loss} and Equation~\ref{gen-loss} in~\ref{append-GAN}. The task is\\then transferred to the long-term DQN using distillation with generated items (Equation~\ref{distill-loss},\\where real items $s_j$ are replaced with items generated by the short-term GAN), while\\previously learnt tasks are retained with pseudo-rehearsal (Equation~\ref{pr-loss}). The long-term GAN\\is then taught to generate data from the new task and previously learnt tasks (Equation~\ref{gan-data},\\where the real items from the new task $s_j$ are replaced with generated items from the\\short-term GAN) using the same loss functions as the short-term GAN.}\\
\hline
$reh\text{-}limit$ & \makecell[l]{This condition is identical to the $reh$ condition above, except that the number of items stored\\and used in rehearsal is limited to the number of uncompressed items (i.e. $600$) which could\\be stored using the same memory allocation size as used by RePR's generative network.}\\
\hline
$reh\text{-}limit\text{-}comp$ & \makecell[l]{This condition is identical to the $reh$ condition above, except that the number of items stored\\and used in rehearsal is limited to the number of compressed items (i.e. $17000$) which could\\be stored using the same memory allocation size as used by RePR's generative network.}\\
\hline
$reh\text{-}extend$ & \makecell[l]{This condition is identical to the $reh$ condition above, except the model and its\\hyper-parameters are set up to learn the extended task sequence (see~\ref{append-extend} for more\\details).}\\
\hline
$RePR\text{-}extend$ & \makecell[l]{This condition is identical to the $RePR$ condition above, except the model and its\\hyper-parameters are set up to learn the extended task sequence (see~\ref{append-extend} for more\\details).}\\
\hline
$reh\text{-}extend\text{-}policy$ & \makecell[l]{This condition is identical to the $reh$ condition above, except the model and its\\hyper-parameters are set up to learn the extended task sequence (see~\ref{append-extend} for more\\details). Furthermore, the policy is transferred and retained in the long-term agent, rather\\than Q-values.}\\
\hline
$RePR\text{-}extend\text{-}policy$ & \makecell[l]{This condition is identical to the $RePR$ condition above, except the model and its\\hyper-parameters are set up to learn the extended task sequence (see~\ref{append-extend} for more\\details). Furthermore, the policy is transferred and retained in the long-term agent, rather\\than Q-values.}\\
\hline
\end{tabular}
}
\end{scriptsize}
\end{center}
\end{table*}

Our current research applies pseudo-rehearsal to deep Q-learning so that a DQN can be used to learn multiple Atari $2600$ games\footnote{A brief summary of the Atari $2600$ games learnt in this paper can be found in~\ref{append-games}.} in sequence. All agents select between 18 possible actions representing different combinations of joystick movements and pressing the fire button. Our DQN is based upon~\cite{mnih2015human} with a few minor changes which we found helped the network to learn the individual tasks quicker. The specifics of these changes can be found in~\ref{append-DQN}. Our initial experiments aim to comprehensively evaluate RePR by comparing it to competing methods. In these experiments, conditions are taught Road Runner, Boxing and James Bond. These were chosen as they were three conceptually different games in which a DQN outperforms humans by a wide margin~\cite{mnih2015human}. The tasks were learnt in the order specified above. The final experiment aims to test the capacity of RePR and analyse whether the model fails gracefully or catastrophically when pushed beyond its capacity. In this condition we extend the sequence learnt with the Atari games Pong, Atlantis and Qbert. Due to the exceptionally long computational time required for running conditions in this experiment, only RePR and rehearsal methods were tested with a single seed on this extended sequence. The procedure for this extended experiment is identical to the procedure described below for the initial experiments, except for two differences. Firstly, learning is slowed down by reducing $\alpha$ and doubling the LTM system's training time. Secondly, a larger DQN and GAN is used to ensure the initial size of the networks is large enough to learn some of the extended sequence of tasks\footnote{Importantly, even when using larger networks, RePR is still pushed beyond its capacity as the results from this experiment clearly demonstrate forgetting.}. Specific details for the extended experiment can be found in~\ref{append-extend}. The details of the experimental conditions used in this paper can be found in Table~\ref{exp-conds}.

The architecture of the DQNs is kept consistent across all experimental conditions that train on the same sequence of tasks. Details of the network architectures (including the GAN) and training parameters used throughout the experiments can be found in~\ref{append-details}. A hyper-parameter search was used in conditions that used a variant of EWC. Details of this search can also be found in this appendix along with other specific implementation details for these conditions. The online-EWC implementation does not include connections from the LTM system to the STM system which try and encourage weight sharing when learning the new task. This was not included because authors of the Progress and Compress method found online-EWC alone was competitive with the Progress and Compress method (which included these connections) and it kept the architecture of the agents' dual memory system consistent with other conditions.

In some of the conditions listed in Table~\ref{exp-conds}, the policy is transferred and retained in the long-term agent, rather than Q-values. Transfer is achieved by giving samples from the current task to the short-term agent and one-hot encoding the action with the largest associated Q-value. This one-hot encoding is then taught to the long-term agent using the cross-entropy loss function. Similarly, when the policy is being retained in RePR's long-term agent, generated samples are given to the previous long-term agent and the softmax values produced are then retained using the cross-entropy loss function. More specific details on learning and retaining the policy (including loss functions) can be found in~\ref{append-policy}.

Each game was learnt by the STM system for $20$ million frames and then taught to the LTM system for $20$m frames. The only exception was for the first long-term DQN which had the short-term DQN's weights copied directly over to it. This means that our experimental conditions differ only after the second task (Boxing) was introduced. The GAN had its discriminator and generator loss function alternatively optimised for $200{,}000$ steps.

When pseudo-rehearsal was applied to the long-term DQN agent or GAN, pseudo-items were drawn from a temporary array of $250{,}000$ states generated by the previous GAN. The final weights for the short-term DQN are those that produce the largest average score over $250{,}000$ observed frames. The final weights for the long-term DQN are those that produced the lowest error over $250{,}000$ observed frames. In the initial experiments $\alpha=0.55$. However, we have also tested RePR with $\alpha=0.35$ and $\alpha=0.75$, both of which produced very similar results, with the final agent performing at least equivalently to the original DQN's results~\cite{mnih2015human} for all tasks.

Our evaluation procedure is similar to~\cite{mnih2015human} in that our network plays each task for $30$ episodes and an episode terminates when all lives are lost. Actions are selected from the network using an $\varepsilon$-greedy policy with $\varepsilon=0.05$. Final network results are also reported using this procedure and standard deviations are calculated over these $30$ episodes. Unless stated otherwise, each condition is trained three times using the same set of seeds between conditions and all reported results are averaged across these seeds.

\section{Results}\label{results}\label{results}

\subsection{RePR Performance on CF}\label{results-main}
The first experiment investigates how well RePR compares to a lower and upper baseline. The $no\text{-}reh$ condition is the lower baseline because it does not contain a component to assist in retaining the previously learnt tasks. The $reh$ condition is the upper baseline for RePR because it rehearses real items from previously learnt tasks and thus, demonstrates how RePR would perform if its GAN could perfectly generate states from previous tasks to rehearse alongside learning the new task\footnote{The $reh$ condition is not doing typical rehearsal although the difference is subtle. It relearns previous tasks using targets produced by the previous network (as in RePR), rather than targets on which the original long-term DQN was taught.}.

\begin{figure}[p]
\vskip 0.2in
\begin{center}
\centerline{\includegraphics[width=\columnwidth,height=0.85\textheight,keepaspectratio]{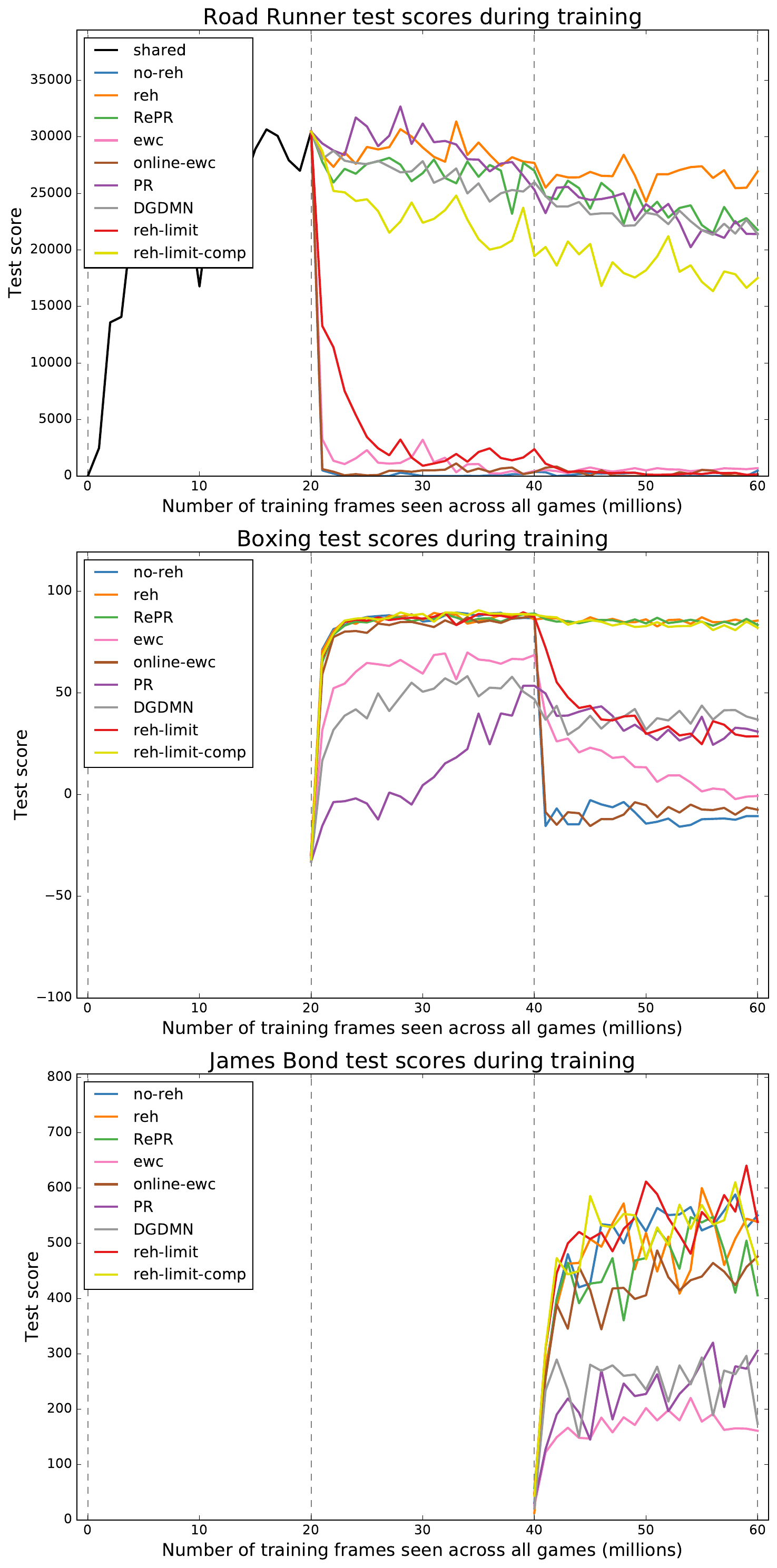}}
\caption{Results of our RePR model compared to the $no\text{-}reh$, $reh$, $ewc$, $online\text{-}ewc$, $PR$, $DGDMN$, $reh\text{-}limit$ and $reh\text{-}limit\text{-}comp$ conditions. After every $1$ million observable training frames, the long-term agent is evaluated on the current task and all previously learnt tasks. Task switches occur at the dashed lines, in the order Road Runner, Boxing and then James Bond.}
\label{main-results}
\end{center}
\vskip -0.2in
\end{figure}

The results of RePR can be found in Fig.~\ref{main-results}, alongside other conditions' results. All of the mentioned conditions outperform the $no\text{-}reh$ condition which severely forgets previous tasks. RePR was found to perform very closely to the $reh$ condition, besides a slight degradation of performance on Road Runner, which was likely due to the GAN performing pseudo-rehearsal to retain states representative of Road Runner. These results suggest that RePR can prevent CF without any need for extra task specific parameters or directly storing examples from previously learnt tasks.

We also investigated whether a similar set of weights are important to the RePR agent's output on all of the learnt tasks or whether the network learns the tasks by dedicating certain weights as important to each individual task. When observing the overlap in the network's Fisher information matrices for each of the games (see~\ref{append-overlap} for implementation details and specific results), we found that the network did share weights between tasks, with similar tasks sharing a larger proportion of important weights. Overall, these positive results show that RePR is a useful strategy for overcoming CF.

\subsection{Quality of Generated Items}\label{results-quality}

\begin{figure}[t]
\vskip 0.2in
\begin{center}
\centerline{\includegraphics[width=\columnwidth]{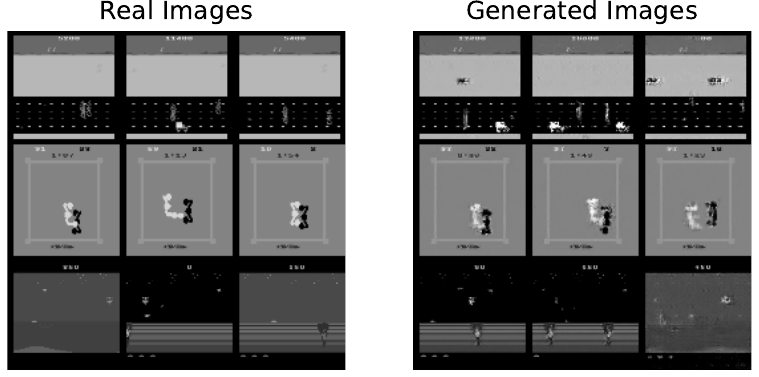}}
\caption{Images drawn from previous tasks' experience replays (real) and images generated from a GAN sequentially taught to produce sequences from Road Runner, Boxing and then James Bond. Images shown are the first image of each four frame sequence. Each row contains images from one of the three tasks.}
\label{real-fake-ims}
\end{center}
\vskip -0.2in
\end{figure}

In RePR, all pseudo-items generated by the GAN are used in pseudo-rehearsal. Furthermore, we have found that balancing the proportion of new items and pseudo-items by the number of tasks learnt (with Equation~\ref{gan-data}) is sufficient to ensure the GAN generates items evenly from each of the tasks. Fig.~\ref{real-fake-ims} shows some GAN generated images after learning all three tasks, alongside real images from the games. This figure shows that although the GAN is successful at generating images similar to the previous games there are clear visual differences between them.

A further experiment was conducted to analytically assess the quality of pseudo-items. This experiment uses three networks. The first is a DQN trained on a task (or sequence of tasks) and is the teacher network. The second is a GAN trained on the same task/s as the first network. The final network is a freshly initialised DQN and is the student network. The student network is taught how to play the task/s by the teacher network (using the distillation loss function in Equation~\ref{distill-loss}). The data used to teach the task/s is either real items from the task/s ($real$) or pseudo-items generated by the GAN ($gan$). Therefore, the score which the student network can attain, after training with real or generated data, reflects the quality of the training data. The student network was either taught to play one of the games: Road Runner ($road$), Boxing ($boxing$) or James Bond ($james$), or taught to play all three of these games at once ($3T$).

\begin{table*}[t]
\caption{Final long-term network scores (and standard deviations) attained after training a student DQN using either real or generated items. Results are collected using a single consistent seed with the average scores and standard deviations calculated by testing the final network on $30$ episodes.}
\label{scratch}
\vskip 0.15in
\begin{center}
\begin{scriptsize}
\begin{sc}
\makebox[\linewidth]{
\begin{tabular}{c|c|c|c}
\hline
Condition & Road Runner & Boxing & James Bond\\
\hline
$real\_road$ & $27133$ ($\pm5772$) & &\\
$gan\_road$ & $15343$ ($\pm5885$) & &\\
$real\_boxing$ & & $84$ ($\pm17$) &\\
$gan\_boxing$ & & $70$ ($\pm21$) &\\
$real\_james$ & & & $482$ ($\pm139$)\\
$gan\_james$ & & & $247$ ($\pm98$)\\
$real\_3T$ & $23660$ ($\pm5112$) & $79$ ($\pm20$) & $342$ ($\pm178$)\\
$gan\_3T$ & $747$ ($\pm739$) & $-6$ ($\pm12$) & $142$ ($\pm100$)\\
\hline
\end{tabular}
}
\end{sc}
\end{scriptsize}
\end{center}
\vskip -0.1in
\end{table*}

Table~\ref{scratch} shows a clear difference between the quality of real and generated items. When learning a single task with items generated by the GAN, the student network cannot learn the task to the same score as it can with real items. When the GAN has been taught multiple tasks, the $gan\_3T$ condition shows that the quality of generated items is severely lower than real items and cannot be used to learn all three tasks to a reasonable standard. This can be considered a positive result for RePR as it demonstrates that pseudo-items can still be used to effectively prevent CF even when the pseudo-items are considerably poorer quality than real items.

\subsection{RePR Versus EWC}

We further investigate the effectiveness of RePR by comparing its performance to the leading EWC variants. The results of both the EWC conditions are also included in Fig.~\ref{main-results}. These results clearly show that RePR outperforms both EWC and online-EWC under these conditions. We find that EWC retains past experiences better than online-EWC and due to this, online-EWC was more effective at learning the new task.

The poor results displayed by the EWC variants contrast substantially from those originally reported by authors~\cite{kirkpatrick2017overcoming,schwarz2018progress}. This can be explained by the differences in our training scheme. More specifically, we do not allow tasks to be revisited, whereas both EWC and Progress and Compress visited tasks several times. Furthermore, we do not allow the networks we tested to grow in capacity when a new task is learnt, whereas the training scheme in~\cite{kirkpatrick2017overcoming} allowed EWC to have two task specific weights per neuron. Finally, the networks we have tested so far are retaining Q-values in their LTM system, whereas Progress and Compress~\cite{schwarz2018progress} retained only the policy of an Actor-Critic network in its LTM system.

In~\ref{append-EWC}, we test RePR and the EWC variants on a different training scheme where only the policy is retained in the LTM system and the total training time of the LTM system is reduced so that tasks need only to be retained for a shorter period of time. Under this different training scheme, the EWC variants perform comparatively to RePR and thus, the training scheme less comprehensively explores the capabilities of the models.

\subsection{Further Evaluating RePR}

In this section, RePR is further evaluated through a number of conditions. Firstly, the $PR$ condition represents an ablation study investigating the importance of the dual memory system. Consequently, the $PR$ condition is identical to the $RePR$ condition without a dual memory system. In Fig.~\ref{main-results}, the $PR$ condition demonstrates poorer results compared to the $RePR$ condition along with slower convergence times for learning the new task. This shows that combining pseudo-rehearsal with a dual memory model, as we have done in RePR, is beneficial for learning the new task.

The $DGDMN$ condition represents an implementation of the DGDMN. This network was proposed for sequential image classification and therefore, significant changes were necessary to allow the model to learn reinforcement tasks. These changes made our DGDMN implementation similar to RePR, except for the presence of a separate GAN in the STM system which DGDMN teaches to generate data representative of the new task. This GAN generates the data used to train the LTM system on the new task. The $DGDMN$ condition also demonstrates poorer results compared to the $RePR$ condition. The most evident difference between these conditions was DGDMN's inability to completely transfer new tasks to its long-term DQN using items generated by its STM system. More specifically, the DGDMN's long-term DQN could learn the new tasks Boxing and James Bond to approximately half the performance of RePR. To be consistent with the other conditions tested, the STM system's networks were copied to the LTM system after learning the first task. Because of this, the LTM system learnt the first task with real data (not data generated from the short-term GAN) and thus, did not struggle to learn the first task Road Runner.

The memory efficiency of RePR is investigated with the $reh\text{-}limit$ and $reh\text{-}limit\text{-}comp$ conditions. These conditions only store a small number of either uncompressed or compressed real items for rehearsal, limited by the memory allocation size of RePR's generative network. The $reh\text{-}limit$ condition shows substantially more forgetting compared to RePR. On Road Runner, the $reh\text{-}limit$ condition quickly forgets everything it has learnt about the task and thus, performs similarly to the $no\text{-}reh$ condition, which makes no effort to retain the task. On Boxing, forgetting causes the $reh\text{-}limit$ condition to retain roughly half of its performance on the task. The $reh\text{-}limit\text{-}comp$ condition only displays forgetting on Road Runner, where, compared to RePR, the condition retains noticeably less knowledge of Road Runner throughout the learning sequence.

\begin{table*}[t]
\caption{Final long-term network scores for each of the conditions, along with their storage requirements. The final three rows contain the scores which the same DQN can attain after training solely on the specified task. Results are collected using three consistent seeds with the average scores and standard deviations calculated by testing each of the seed's final networks on $30$ episodes.}
\label{score-table}
\vskip 0.15in
\begin{center}
\begin{scriptsize}
\begin{sc}
\makebox[\linewidth]{
\begin{tabular}{c|c|c|c|c}
\hline
Condition & \multicolumn{3}{c|}{Final network's average score (std)} & Memory space*\\
 & Road Runner & Boxing & James Bond & \\
\hline
$no\text{-}reh$ & $0$ ($\pm0$) & $-13$ ($\pm9$) & $619$ ($\pm165$) & $0.007$ GB\\
$reh$ & $26486$ ($\pm6245$) & $85$ ($\pm10$) & $548$ ($\pm156$) & $7.063$ GB\\
$RePR$ & $22042$ ($\pm5375$) & $82$ ($\pm12$) & $468$ ($\pm155$) & $0.023$ GB\\
$ewc$ & $581$ ($\pm525$) & $-2$ ($\pm10$) & $162$ ($\pm97$) & $0.042$ GB\\
$online\text{-}ewc$ & $130$ ($\pm159$) & $-10$ ($\pm10$) & $381$ ($\pm165$) & $0.014$ GB\\
$PR$ & $21804$ ($\pm6042$) & $33$ ($\pm17$) & $342$ ($\pm99$) & $0.023$ GB\\
$DGDMN$ & $22841$ ($\pm4211$) & $36$ ($\pm24$) & $184$ ($\pm108$) & $0.023$ GB\\
$reh\text{-}limit$ & $151$ ($\pm329$) & $29$ ($\pm26$) & $598$ ($\pm246$) & $0.024$ GB\\
$reh\text{-}limit\text{-}comp$ & $16348$ ($\pm6980$) & $80$ ($\pm14$) & $511$ ($\pm178$) & $0.024$ GB\\
\hline
Road Runner & $29463$ ($\pm7864$) & & & n/a\\
Boxing & & $88$ ($\pm8$) & & n/a\\
James Bond & & & $645$ ($\pm161$) & n/a\\

\hline
\multicolumn{5}{l}{*Memory allocation size required for long-term storage.}
\end{tabular}
}
\end{sc}
\end{scriptsize}
\end{center}
\vskip -0.1in
\end{table*}

The scores for the final long-term networks\footnote{For the $PR$ condition, as no dual memory system is used, we refer to the DQN which learns new tasks while retaining previous tasks as the final long-term network.} in each of the conditions are shown in Table~\ref{score-table} along with the long-term storage space required by the models to be able to continue learning. Additionally, the table includes the scores which can be attained by training three DQNs individually on each of the tasks. Similar to previous results, this table shows that the $reh$ and $RePR$ conditions are the most successful at learning and retaining this sequence of tasks, with $reh$ achieving on average $90\%$ of the scores that could be attained from individually learning the tasks and $RePR$ achieving $80\%$. The scores attained by RePR are found to be well above human expert performance levels ($7845$, $4$, $407$)~\cite{mnih2015human}. The table also shows that the long-term memory allocation used by the $RePR$ condition was smaller than the $reh$ condition by orders of magnitude\footnote{When the $reh$ condition is storing compressed items, the size of its long-term memory allocation is still an order of magnitude greater than $RePR$ at approximately $0.250$ GB.}. Although, we did not attempt to optimise this size for either of the conditions, the $reh\text{-}limit$ and $reh\text{-}limit\text{-}comp$ conditions show that RePR outperforms rehearsal when rehearsal is constrained to approximately the same memory size as the $RePR$ condition.

\begin{figure}[p]
\vskip 0.2in
\begin{center}
\centerline{\includegraphics[width=\columnwidth,height=0.85\textheight,keepaspectratio]{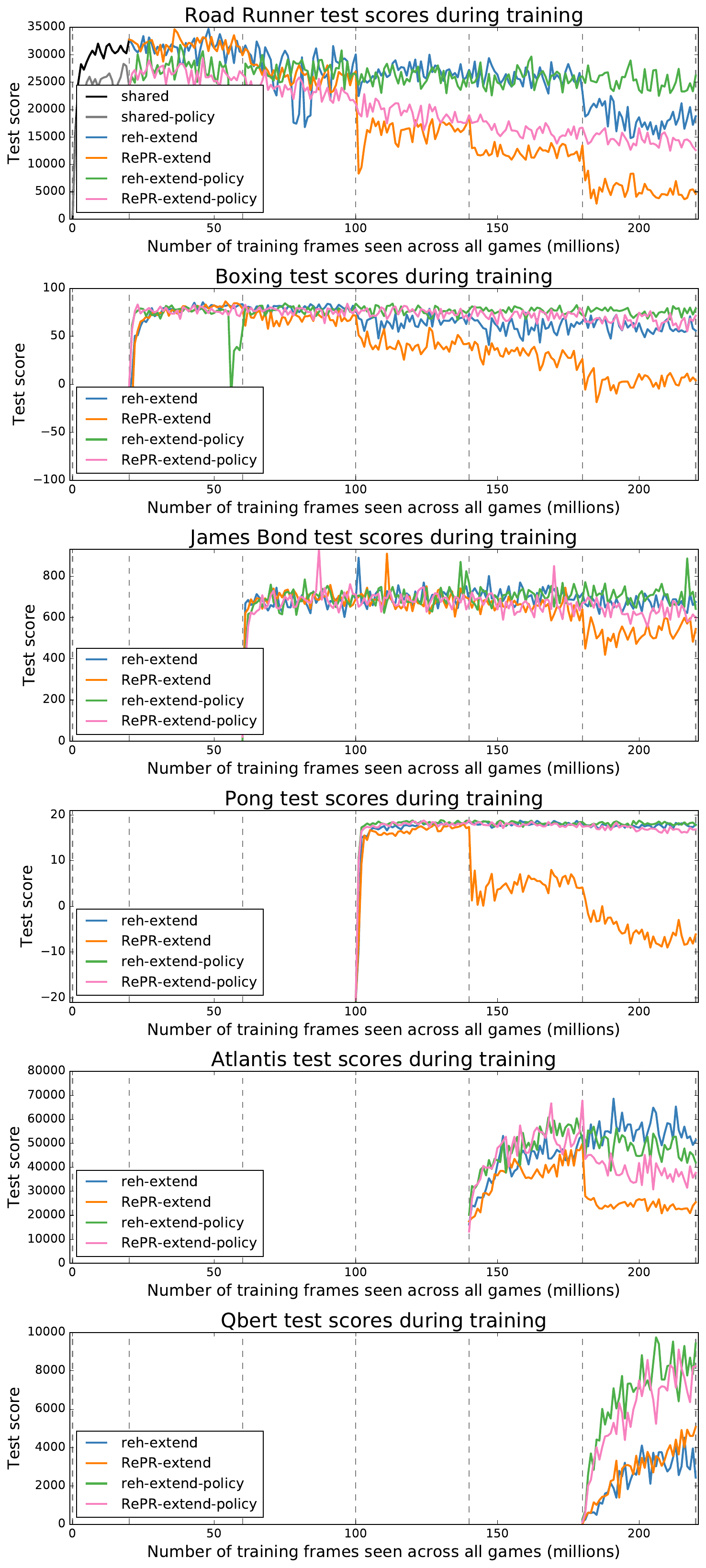}}
\caption{Results of the $reh\text{-}extend$, $RePR\text{-}extend$, $reh\text{-}extend\text{-}policy$ and $RePR\text{-}extend\text{-}policy$ conditions on an extended sequence of tasks. After every $1$ million observable training frames, the long-term agent is evaluated on the current task and all previously learnt tasks. Task switches occur at the dashed lines, in the order Road Runner, Boxing, James Bond, Pong, Atlantis and then Qbert. Results were produced using a single seed.}
\label{extend-results}
\end{center}
\vskip -0.2in
\end{figure}

Finally, Fig.~\ref{extend-results} displays the results for the final experiment which compares RePR to rehearsal in the extended sequence of Atari $2600$ tasks. Overall, the $reh\text{-}extend$, $reh\text{-}extend\text{-}policy$ and $RePR\text{-}extend\text{-}policy$ conditions retained considerable knowledge of previous tasks, with $reh\text{-}extend\text{-}policy$ retaining noticeably more knowledge of Road Runner. The $RePR\text{-}extend$ condition underperformed compared to the other conditions, showing mainly a gradual decline in performance of previously learnt tasks.

\section{Discussion}

Our experiments have demonstrated RePR to be an effective solution to CF when sequentially learning multiple tasks. To our knowledge, pseudo-rehearsal has not been used until now to successfully prevent CF on complex reinforcement learning tasks. RePR has advantages over popular weight constraint methods, such as EWC, because it does not constrain the network to retain similar weights when learning a new task. This allows the internal layers of the network to change according to new knowledge, giving the model the freedom to restructure itself when incorporating new information. Experimentally, we have verified that RePR does outperform these state-of-the-art EWC methods on a sequential learning task.

\subsection{Dual Memory}

The $PR$ condition omits using the dual memory system to analyse the system's importance. This results in the condition trying to learn non-stationary target Q-values (for the new task) through reinforcement learning, while also retaining previous tasks through pseudo-rehearsal. The results for the $PR$ condition showed that convergence times were longer and the final performance reached was lower on new tasks compared to $RePR$. Therefore, this demonstrates that omitting the dual memory system increases the interference between new knowledge and knowledge of previous tasks. In~\ref{append-DM} the extent of this interference is explored and it is found that to learn Boxing to a similar standard to RePR, while also omitting the dual memory system, the deep Q-learning loss function must be weighted substantially higher than the pseudo-rehearsal loss function, such that the model considerably suffers from CF. Overall, these results are strong evidence that the dual memory system is beneficial in sequential reinforcement learning tasks. This is particularly interesting because the authors of Progress and Compress~\cite{schwarz2018progress} did not find clear evidence that their algorithm benefited from a similar dual memory system in their experimental conditions and additionally, the dual memory system in RePR showed more substantial benefits than the dual memory system used by DGDMN in image classification with the same number of tasks~\cite{kamra2017deep}.

Our results show that the dual memory system used by DGDMN performs substantially worse than RePR in reinforcement learning. This is because DGDMN relies on short-term generative networks to provide the data to teach the LTM system. When these tasks are difficult for the generative network to learn, this data is not accurate enough to effectively teach new tasks to the LTM system. Although this was not an issue for the DGDMN in image classification~\cite{kamra2017deep}, in Section~\ref{results-quality} we show that, in these complex reinforcement tasks, the GAN does struggle such that it is much more effective to teach a new task to a freshly initialised DQN using real data than generated data. However, our results also show that generated items that are not of high enough quality for STM to LTM transfer, can be used effectively in pseudo-rehearsal.

To overcome the limited quality of generative items, RePR's dual memory system assumes real data from the most recent task can be retained in an experience replay so that it can be used to successfully teach the LTM system. We do not believe this assumption is unrealistic because the experience replay is a temporary buffer, with a fixed size, and once the most recent task has been consolidated into LTM, it is cleared to make room for data from the next task.

\subsection{Scalability}

One main advantage of RePR compared to many other continual learning algorithms is that it is very scalable. Applying pseudo-rehearsal to both the agent and generative model means that the network does not need to use any task specific weights to accomplish continual learning and thus, the model's memory requirements do not grow as the number of tasks increases. RePR's generative network also means that it does not retain knowledge by storing raw data from previous tasks. This is advantageous in situations where storing raw data is not allowed because of e.g. privacy reasons, biological plausibility, etc. Furthermore, results show RePR outperforms rehearsal when the number of rehearsal items are limited by the storage space required for RePR's generative network ($reh\text{-}limit$ and $reh\text{-}limit\text{-}comp$). Therefore, RePR can also prevent CF more effectively than rehearsal when memory is limited.

In theory, RePR could be used to learn any number of tasks as long as the agent's network (e.g. DQN) and GAN have the capacity to successfully learn the collection and generate states that are representative of previously learnt tasks. However, in practice the capacity of RePR's components could be exceeded and we explored how this affects RePR by investigating it on an extended sequence of tasks. In this test, RePR demonstrated a modest amount of forgetting when required to retain only the policy in its LTM system. However, when RePR is required to retain Q-values in its LTM system, it shows substantial forgetting compared to rehearsal. In this case, RePR's forgetting was generally gradual, such that learning the new tasks only disrupts retention after millions of training frames. However, the $reh\text{-}extend$ condition does not notably suffer from this gradual forgetting. The only difference between these conditions is that the $reh\text{-}extend$ condition uses real items to rehearse previous tasks. This is important as it identifies that it is the GAN which is struggling to learn the extended sequence of tasks, resulting in the observed forgetting. Therefore, future research improving GANs' capacity and stability (vanishing gradients and mode collapse~\cite{li2018limitations}) will directly improve RePR's ability to prevent CF in these challenging conditions.

\subsection{Limitations and Future Research}

One limitation of rehearsal methods (including RePR) is that they become infeasible when the number of tasks goes to infinity. This is because the items rehearsed in each training iteration will only cover a very small selection of previously seen tasks. However, research in sweep rehearsal~\cite{silver2015consolidation,robins1995catastrophic} shows that rehearsal methods can still be beneficial when rehearsing a comparably small number of items per task, suggesting that this limitation will only become severe in long task sequences. Another limitation of our model is that it currently assumes the agent knows when a task switch occurs. However, in some use cases this might not be true and thus, a task detection method would need to be combined with RePR.

In these experiments we use the same task sequence for all conditions which could bias the results either positively or negatively.  However, there is work~\cite{poirier2005effect} which suggests that for a given set of tasks, the mean performance of the final model is not affected by the order in which those tasks are learnt.

Currently, our RePR model has been designed to prevent the CF that occurs while learning a sequence of tasks, without using real data. However, our implementation still stores a limited amount of real data from the current task in an experience replay. This data is used to prevent the CF that occurs while learning a single task. We believe this was acceptable as our goal in this research was to prevent the CF that occurs from sequential task learning (without using real data) and consequently, this real data was never used to prevent this form of CF. In future work we wish to extend our model by further modifying pseudo-rehearsal so that it can also prevent the CF that occurs while learning a single reinforcement task.

In this paper, we chose to apply our RePR model to DQNs. However, this can be easily extended to other state-of-the-art deep reinforcement learning algorithms, such as actor-critic networks, by adding a similar constraint to the loss function. We chose DQNs for this research because their Q-values contain both policy and value information whereas Actor-Critics produce policy and value information separately. Although our research suggests that it is easier for the LTM system to retain solely the policy, the policy does not contain information about the expected discounted rewards (i.e. the value) associated with each state which is necessary to continue learning an already seen task. Future research could investigate whether the expected discounted rewards can be quickly relearnt with access to the retained policy, or whether it should also be retained by the LTM system to continue reinforcement learning without disruption.

\section{Conclusion}

In conclusion, pseudo-rehearsal can be used with deep reinforcement learning methods to achieve continual learning. We have shown that our RePR model can be used to sequentially learn a number of complex reinforcement tasks, without scaling in complexity as the number of tasks increases and without revisiting or storing raw data from past tasks. Pseudo-rehearsal has major benefits over weight constraint methods as it is less restrictive on the network and this is supported by our experimental evidence. We also found compelling evidence that the addition of our dual memory system is necessary for continual reinforcement learning to be effective. As the power of generative models increases, it will have a direct impact on what can be achieved with RePR and our goal of having an agent which can continuously learn in its environment, without being challenged by CF.

\appendix

\section{Details on the Atari $2600$ Games}\label{append-games}
Road Runner is a game where the agent must outrun another character by moving toward the left of the screen while collecting items and avoiding obstacles. To achieve high performance the agent must also learn to lead its opponent into certain obstacles to slow it down. Boxing is a game where the agent must learn to move its character around a 2D boxing ring and throw punches aimed at the face of the opponent to score points, while also avoiding taking punches to the face. James Bond has the agent learn to control a vehicle, while avoiding obstacles and shooting various objects. In Pong, the agent learns to hit a ball back to its opponent by moving a paddle. A point is scored by the agent when the opponent does not successfully hit the ball back and the opponent scores a point when the agent does not successfully hit back the ball. For Atlantis, the agent learns to control three stationary cannons and must shoot down enemy ships moving horizontally across the sky. After a ship passes 4 times it will take out one of the city's bases, starting with the central cannon. The agent scores points for shooting down ships and loses once all bases have been destroyed. Finally, in Qbert the agent learns to control a character which jumps diagonally around a pyramid of cubes, changing the cubes' colours. Once all the cubes have been changed to a particular colour the level is cleared. In this game, the agent must also learn to avoid various enemies.

\section{Further Implementation Details}\label{append-details}
\subsection{DQN}\label{append-DQN}

The main difference between our DQN and~\cite{mnih2015human} is that we used TensorFlow's RMSProp optimiser (without centering) with global norm gradient clipping compared to the original paper's RMSProp optimiser which clipped gradients between $[-1, 1]$. Our network architecture remained the same. However, our biases were set to $0.01$ and weights were initialised with $\mathcal{N}(0, 0.01)$, where all values that were more than two standard deviations from the mean were re-drawn. The remaining changes were to the hyper-parameters of the learning algorithm which can be seen in bold in Table~\ref{dqn-hyp}. The architecture of our network can be found in Table~\ref{dqn-arch}, where all layers use the ReLU activation function except the last linear layer. 

\begin{table*}[t]
\caption{DQN hyper-parameters.}
\label{dqn-hyp}
\renewcommand{\arraystretch}{1.5}
\begin{center}
\begin{scriptsize}
\makebox[\linewidth]{
\begin{tabular}{c | c | l}
\hline
Hyper-parameter & Value & Description\\
\hline
mini-batch size & 32 & \makecell[l]{Number of examples drawn for calculating the stochastic gradient\\descent update.}\\
replay memory size & \textbf{200{,}000} & \makecell[l]{Number of frames in experience replay which samples from the\\current game are drawn from.}\\
history length & 4 & Number of recent frames given to the agent as an input sequence.\\
target network update frequency & \textbf{5{,}000} & \makecell[l]{Number of frames which are observed from the environment before\\the target network is updated.}\\
discount factor & 0.99 & Discount factor ($\gamma$) for each future reward.\\
action repeat & 4 & \makecell[l]{Number of times the agent's selected action is repeated before\\another frame is observed.}\\
update frequency & 4 & \makecell[l]{Frequency of observed frames which updates to the current network\\occur on.}\\
learning rate & 0.00025 & Learning rate used by Tensorflow's RMSProp optimiser. \\
momentum & 0.0 & Momentum used by Tensorflow's RMSProp optimiser. \\
decay & \textbf{0.99} & Decay used by Tensorflow's RMSProp optimiser. \\
epsilon & $\pmb{1e^{-6}}$ & Epsilon used by Tensorflow's RMSProp optimiser. \\
initial exploration & 1.0 & Initial $\varepsilon$-greedy exploration rate.\\
final exploration & 0.1 & Final $\varepsilon$-greedy exploration rate.\\
final exploration frame & 1{,}000{,}000 & \makecell[l]{Number of frames seen by the agent before the linear decay of the\\exploration rate reaches its final value.}\\
replay start size & 50{,}000 & \makecell[l]{The number of frames which the experience replay is initially filled\\with (using a uniform random policy).}\\
no-op max & 30 & \makecell[l]{Maximum number of ``do nothing" actions performed at the start of\\an episode ($U[1,\text{no-op max}]$).}\\
\hline
\end{tabular}
}
\end{scriptsize}
\end{center}
\end{table*}

\begin{table}[t]
\caption{DQN architecture used in all experiments except the extended task sequence experiment, where CONV is a convolutional layer and FC is a fully connected layer.}
\label{dqn-arch}
\begin{center}
\begin{scriptsize}
\makebox[\linewidth]{
\begin{tabular}{cccc}
\hline
\multicolumn{4}{c}{DQN}\\
\hline
\multicolumn{4}{c}{Input: $4 \times 84 \times 84$}\\
layer & \# units/filters & filter shape & filter stride\\
\hline
CONV & $32$ & $8 \times 8$ & $4 \times 4$\\
CONV & $64$ & $4 \times 4$ & $2 \times 2$\\
CONV & $64$ & $3 \times 3$ & $1 \times 1$\\
FC & $512$\\
FC & $18$\\
\hline
\end{tabular}
}
\end{scriptsize}
\end{center}
\end{table}

\subsection{GAN}\label{append-GAN}

The GAN is trained with the WGAN-GP loss function~\cite{gulrajani2017improved} with a drift term~\cite{karras2017progressive}. The drift term is applied to the discriminator's output for real and fake inputs, stopping the output from drifting too far away from zero. More specifically, the loss functions used for updating the discriminator ($L_{disc}$) and generator ($L_{gen}$) are:
\newcommand\norm[1]{\left\lVert#1\right\rVert}
\begin{equation}
\label{disc-loss}
\begin{aligned}
L_{disc} = D(\widetilde{x}; \phi) - D(x; \phi) + \lambda(\norm{\nabla_{\hat{x}} D(\hat{x}; \phi)}_2-1)^2 \\ + \epsilon_{drift} D(x; \phi)^2 + \epsilon_{drift} D(\widetilde{x}; \phi)^2,
\end{aligned}
\end{equation} 
\begin{equation}
\label{gen-loss}
\begin{aligned}
L_{gen} = -D(\widetilde{x}; \phi),
\end{aligned}
\end{equation} where $D$ and $G$ are the discriminator and generator networks with the parameters $\phi$ and $\varphi$. $x$ is an input item drawn from either the current task's experience replay or the previous long-term GAN (as specified in Equation~\ref{gan-data}). $\widetilde{x}$ is an item produced by the current generative model ($\widetilde{x} = G(z; \varphi)$) and $\hat{x} = \epsilon x + (1 - \epsilon)\widetilde{x}$. $\epsilon$ is a random number $\epsilon \sim U(0,1)$, $z$ is an array of latent variables $z = U(-1, 1)$, $\lambda = 10$ and $\epsilon_{drift} = 1e^{-6}$. The discriminator and generator networks' weights are updated on alternating steps using their corresponding loss function.

\begin{table*}[t]
\caption{GAN architecture used in all experiments except the extended task sequence experiment, where FC is a fully connected layer, DECONV is a deconvolutional layer and CONV is a convolutional layer.}
\label{gan-arch}
\begin{center}
\begin{scriptsize}
\makebox[\linewidth]{
\begin{tabular}{cccc | cccc}
\hline
\multicolumn{4}{c|}{Generator} & \multicolumn{4}{c}{Discriminator}\\
\hline
\multicolumn{4}{c|}{Input: $100$ latent variables} & \multicolumn{4}{c}{Input: $4 \times 84 \times 84$}\\
layer & \# units/filters & filter shape & filter stride & layer & \# units/filters & filter shape & filter stride\\
\hline
FC & $256 \times 7 \times 7$ & & & CONV & $64$ & $5 \times 5$ & $3 \times 3$\\
DECONV & $256$ & $5 \times 5$ & $3 \times 3$ & CONV & $128$ & $5 \times 5$ & $2 \times 2$\\
DECONV & $128$ & $5 \times 5$ & $2 \times 2$ & CONV & $256$ & $5 \times 5$ & $2 \times 2$\\
DECONV & $64$ & $5 \times 5$ & $2 \times 2$ & FC & $1$\\
DECONV & $4$ & $5 \times 5$ & $1 \times 1$\\
\hline
\end{tabular}
}
\end{scriptsize}
\end{center}
\end{table*}

The GAN is trained with the Adam optimiser ($\alpha = 0.001$, $\beta_1 = 0.0$, $\beta_2 = 0.99$ and $\epsilon = 1e^{-8}$ as per~\cite{karras2017progressive}) where the networks are trained with a mini-batch size of 100. The architecture of the networks is illustrated in Table~\ref{gan-arch}. All layers of the discriminator use the Leaky ReLU activation function (with the leakage value set to $0.2$), except the last linear layer. All layers of the generator use batch normalisation ($momentum = 0.9$ and $\epsilon = 1e^{-5}$) and the ReLU activation function, except the last layer which has no batch normalisation and uses the Tanh activation function. This is to make the generated images' output space the same as the real images which are rescaled between $-1$ and $1$ by applying $f(x) = 2(\frac{x}{255} - 0.5)$ to each raw pixel value. We also decreased the convergence time of our GAN by applying random noise $U(-10, 10)$ to real and generated images before rescaling and giving them to the discriminator.

\subsection{EWC}\label{append-EWC}

The EWC constraint is implemented as per~\cite{kirkpatrick2017overcoming}, where the loss function is amended so that:
\begin{equation}
\label{ltm-ewc-loss}
L_{LTM\_EWC} = \frac 1 N \sum\limits_{j = 1}^{N} {L_D}_j + \frac \lambda 2 {L_{EWC}},
\end{equation}
\begin{equation}
\label{ewc-loss}
L_{EWC} =\sum\limits_i F_i(\theta_i - \theta_i^*)^2,
\end{equation} where $L_D$ is the distillation loss for learning the current task (as specified in Equation~\ref{distill-loss}) and $N$ is the batch-size. $\lambda$ is a scaling factor determining how much importance the constraint should be given, $\theta$ is the current long-term network's parameters, $\theta^*$ is the final long-term network's parameters after learning the previous task and $i$ iterates over each of the parameters in the network. $F_i$ is an approximation of the diagonal elements in a Fisher information matrix, where each element represents the importance each parameter has on the output of the network.

The Fisher information matrix is calculated as in~\cite{kirkpatrick2017overcoming}, by approximating the posterior as a Gaussian distribution with the mean given by the optimal parameters after learning a previous task $\theta_i^*$ and a standard deviation $\beta = 1$. More specifically, the calculation follows~\cite{pascanu2013revisiting}:
\begin{equation}
\label{fisher-eq}
F =\beta^2\mathbb{E}_{s \sim U(D)}[\boldsymbol{J}_y^T \boldsymbol{J}_y],
\end{equation} where an expectation is calculated by uniformly drawing states from the experience replay ($s \sim U(D)$). $\boldsymbol{J}_y$ is the Jacobian matrix $\frac {\partial y} {\partial \theta}$ for the output layer $y$.

When the standard EWC implementation is extended to a third task, a separate penalty is added. This means the current parameters of the network are constrained to be similar to both the parameters after learning the first task and the parameters after further learning the second task.

Online-EWC further extends EWC so that only the previous network's parameters and a single Fisher information matrix is stored. As per~\cite{schwarz2018progress}, this results in the $L_{EWC}$ constraint being replaced by:
\begin{equation}
\label{oewc-loss}
L_{OEWC} =\sum\limits_i F^*_i(\theta_i - \theta_{i,{t-1}}^*)^2,
\end{equation} where the single Fisher information matrix $F^*$ is updated by:
\begin{equation}
\label{oewc-Fupdate}
F^* = \gamma F^*_{t-1} + F_t,
\end{equation} where $\gamma < 1$ is a discount parameter and $t$ represents the index of the current task. In online-EWC, Fisher information matrices are normalised using min-max normalisation so that the tasks' different reward scales do not affect the relative importance of parameters between tasks.

For the $ewc$ condition, we applied a grid search over $\lambda = [50, 100, 150, 200,\\250, \textbf{300}, 350, 400]$ and for our $online\text{-}ewc$ condition we performed a grid search over $\lambda = [25, \textbf{75}, 125, 175]$ and $\gamma = [0.95, \textbf{0.99}]$. The best parameters found during the grid searches are in bold. In all conditions, the Fisher information matrix is calculated by sampling 100 batches from each task. The final network's test scores for each of the tasks were min-max normalised and the network with the best average score was selected. The minimum and maximum is found across all testing episodes played during the learning of the task in the STM system.

\begin{figure}[p]
\vskip 0.2in
\begin{center}
\centerline{\includegraphics[width=\columnwidth,height=0.85\textheight,keepaspectratio]{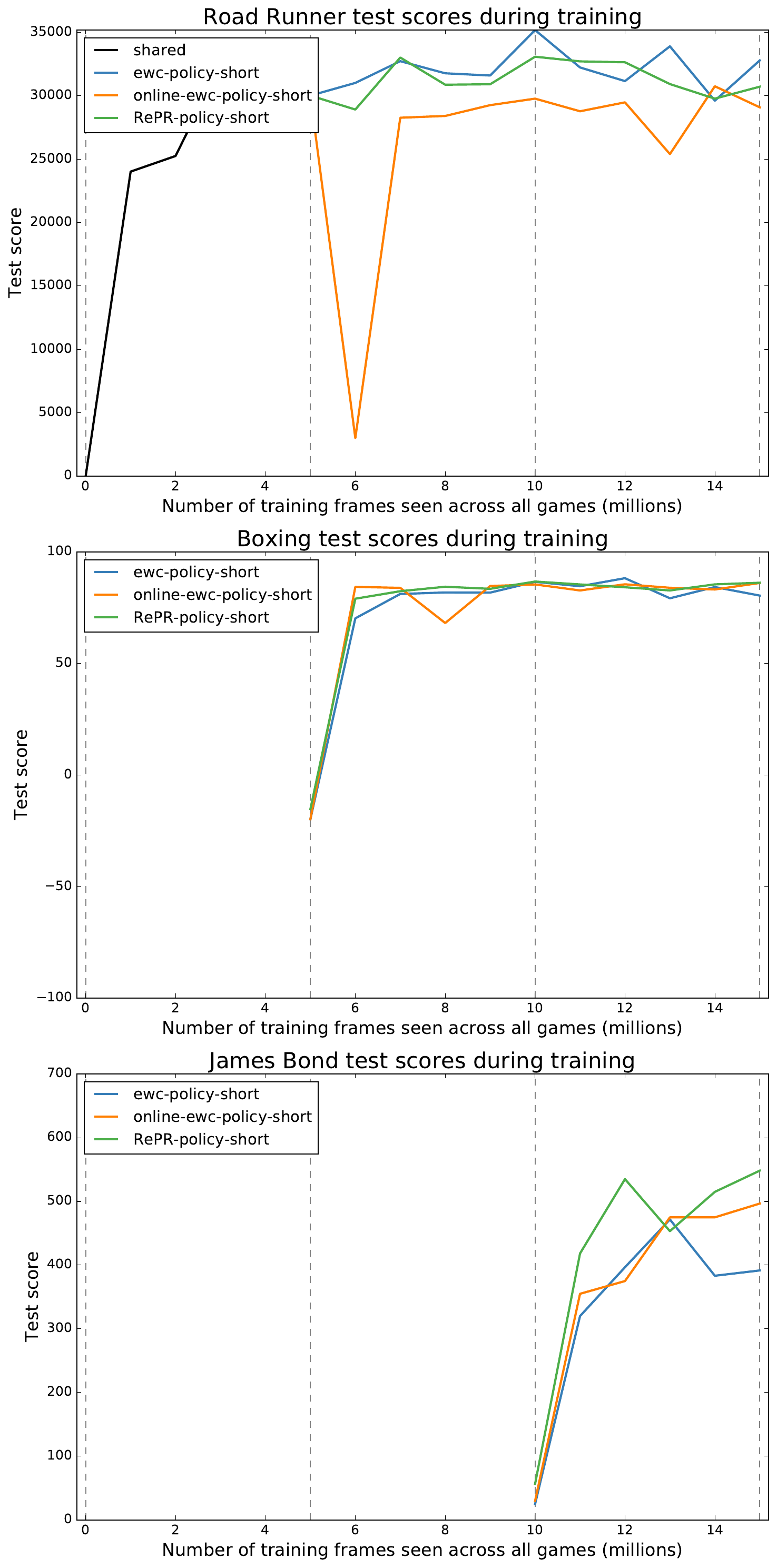}}
\caption{Results of our EWC, online-EWC and RePR implementations tested under a different training scheme. After every $1$ million observable training frames, the long-term agent is evaluated on the current task and all previously learnt tasks. Task switches occur at the dashed lines, in the order Road Runner, Boxing and then James Bond. Results were produced using a single seed.}
\label{policy-only-results}
\end{center}
\vskip -0.2in
\end{figure}

An additional experiment was run to confirm our EWC variants could successfully retain previous task knowledge under a different training scheme. In these conditions, the LTM system retained the agent's policy (taught by minimising the cross-entropy) and new tasks were learnt in the LTM system for 5m frames each. Fig.~\ref{policy-only-results} displays results for the EWC, online-EWC and RePR implementations tested under these conditions ($ewc\text{-}policy\text{-}short$, $online\text{-}ewc\text{-}policy\text{-}short$ and $RePR\text{-}policy\text{-}short$ respectively). All conditions performed similarly and could successfully learn new tasks while retaining knowledge of previous tasks.

\subsection{Transferring and Retaining the Policy in the LTM System}\label{append-policy}
Firstly, the policy is extracted from the short-term DQN by giving samples from the current task to the DQN and calculating the Q-values it has learnt to associate with those samples. For each sample, the action with the largest Q-value is then one-hot encoded. After this policy has been extracted, distillation is used to transfer it to the long-term agent, using the loss function:
\begin{equation}
\label{distill-loss-policy}
{L_{D\_policy}}_j = CE(\pi(s_j;\theta_i), h(Q(s_j;\psi_i))),
\end{equation} where $s_j$ is a state drawn from the current task's experience replay. $\pi$ is the long-term agent which has a softmax output layer and $Q$ is the short-term DQN agent which has a linear output layer. $\theta_i$ is the long-term agent's weights on the current task and $\psi_i$ is the short-term agent's weights after learning the current task. $CE$ is the standard cross-entropy loss function and $h(q)$ is a function that one-hot encodes $q$.

For EWC variants, the long-term agent retains the policy through the same methods it retains Q-values (i.e. using a weight constraint). However, in rehearsal based methods (including RePR) it is necessary to adapt the retention loss function to use cross-entropy. More specifically, the loss function for RePR is changed to:
\begin{equation}
\label{pr-loss-policy}
{L_{PR\_policy}}_j =CE(\pi(\widetilde{s}_j;\theta_i), \pi(\widetilde{s}_j;\theta_{i-1})),
\end{equation} where pseudo-items' inputs $\widetilde{s}_j$ are generated from a GAN and are representative of states in previously learnt games. $\theta_{i-1}$ is the long-term agent's weights after learning the previous task.

\subsection{Extended Task Sequence Experiment}\label{append-extend}
The experiment investigating RePR with an extended task sequence is nearly identical to the first experiment's details from the main text. One of the main differences is that learning is slowed down by initially setting $\alpha$ to $0.05$ and after learning the fourth task reducing this to $0.01$. Consequently, training time is increased to $40$m frames for the long-term DQN. Secondly, both the DQN and generative networks are enlarged by doubling the number of filters and units in hidden layers. More specifically, the convolutional layers in the DQN use $64$, $128$ and $128$ filters respectively and then the fully connected layer (before the output layer) uses $1024$ units. For the GAN, the generative network is enlarged by increasing the first fully connected layer to $512 \times 7 \times 7$ units and increasing the following three deconvolutional layers so that they use $512$, $256$ and $128$ filters respectively. The training time for the GAN is also increased to $400{,}000$ steps.

The $reh\text{-}extend$ and $RePR\text{-}extend$ conditions learn the extended sequence of tasks with either rehearsal or RePR. In these conditions we standardise the short-term DQN's Q-values when they are being taught to the long-term DQN. This was beneficial because it reduced the interference between the games' substantially different reward functions and thus, evened out the importance of retaining each of the games. In the $reh\text{-}extend\text{-}policy$ and $RePR\text{-}extend\text{-}policy$ conditions only the policy was retained in the LTM system. This was achieved by using cross-entropy to learn and retain each game's policy, as described above.

\section{How well does RePR Share Weights?}\label{append-overlap}

To investigate whether an agent's DQN uses similar parameters for determining its output across multiple tasks,~\cite{kirkpatrick2017overcoming} suggest that the degree of overlap between two tasks' Fisher information matrices can be analysed. This Fisher overlap score is bounded between 0 and 1, where a high score represents high overlap and indicates that many of the weights that are important for calculating the desired action in one task are also important in the other task. More specifically, the Fisher overlap is calculated by $1 - d^2$, where:
\begin{equation}
d^2(\hat{F}_1, \hat{F}_2) = \frac 1 2 tr \Big( \hat{F}_1 + \hat{F}_2 - 2(\hat{F}_1\hat{F}_2)^{\frac 1 2} \Big),
\end{equation} given $\hat{F}_1$ and $\hat{F}_2$ are the two tasks' Fisher information matrices which have been normalised so that they each have a unit trace. Fisher information matrices are approximated by Equation~\ref{fisher-eq} using 100 batches of samples drawn from each tasks' experience replay.

\begin{table*}[t]
\caption{Fisher overlap scores between task pairs.}
\label{F-over}
\vskip 0.15in
\begin{center}
\begin{scriptsize}
\begin{sc}
\makebox[\linewidth]{
\begin{tabular}{c|c|c|c}
\hline
Condition & Road Runner \& Boxing & Road Runner \& James Bond & Boxing \& James Bond\\
\hline
$RePR$ & 0.691 & 0.233 & 0.198\\
$RePR\text{-}rev$ & 0.753 & 0.192 & 0.110\\
\hline
\end{tabular}
}
\end{sc}
\end{scriptsize}
\end{center}
\vskip -0.1in
\end{table*}

We compared RePR's Fisher information matrices for each task using the Fisher overlap calculation. When RePR had learnt the tasks in the order Road Runner, Boxing and then James Bond (as in the $RePR$ condition from Section~\ref{results-main}), the Fisher overlap score was high between the first two tasks learnt but relatively low between other task pairs. This suggests that there are more similarities between Road Runner and Boxing than other task pairs. We confirm this by calculating the Fisher overlap for each of the task pairs when the RePR model had successfully learnt the tasks in the reverse order (ie. James Bond, Boxing and then Road Runner). In this case, a higher overlap value remains between Road Runner and Boxing, regardless of the order they were learnt in. This demonstrates that the network attempts to share the computation across a similar set of important weights, where the more similar the tasks are the more effective they are at sharing weights. The precise Fisher overlap values for both of these conditions can be found in Table~\ref{F-over}.

\section{Evaluating the Importance of Dual Memory}\label{append-DM}

To further evaluate the importance of the dual memory system, the $PR$ and $RePR$ conditions from the main text are trained with varying importance values ($\alpha$) on the sequence Road Runner, then Boxing. The $PR$ condition does not use a dual memory system and thus, its $\alpha$ value weights the importance of learning new tasks with deep Q-learning (Equation~\ref{dqn-loss}) vs. retaining previous tasks with pseudo-rehearsal (Equation~\ref{pr-loss}). The $RePR$ condition has a dual memory system and thus, its $\alpha$ value weights the importance of learning new tasks with distillation (Equation~\ref{distill-loss}) vs. retaining previous tasks with pseudo-rehearsal (Equation~\ref{pr-loss}). The aim of this experiment was to investigate the extent to which omitting the dual memory system increases the interference between new knowledge and knowledge of old tasks. This was investigated by finding the $\alpha$ value the $PR$ condition needed to learn Boxing to the same standard as RePR (i.e. to an approximate score of $80$) and what effect these varying $\alpha$ values had on the two models' retention.

\begin{table*}[t]
\caption{Final long-term network scores (and standard deviations) for the $RePR$ and $PR$ models after learning Road Runner and Boxing with varying importance values ($\alpha$). Results are collected using a single consistent seed with the average scores and standard deviations calculated by testing the final network on $30$ episodes.}
\label{lscale}
\begin{center}
\begin{scriptsize}
\makebox[\linewidth]{
\begin{tabular}{c | cc | cc}
\hline
& \multicolumn{2}{c|}{$PR$} & \multicolumn{2}{c}{$RePR$}\\
$\alpha$ & Road Runner & Boxing & Road Runner & Boxing\\
\hline
$0.65$ & $30700$ ($\pm6463)$ & $32$ ($\pm17)$ & $28543$ ($\pm4699)$ & $84$ ($\pm9)$\\
$0.75$ & $30630$ ($\pm5420$) & $32$ ($\pm21$) & $23603$ ($\pm6124$) & $85$ ($\pm8$)\\
$0.85$ & $26310$ ($\pm8257$) & $46$ ($\pm14$) & $13977$ ($\pm8814$) & $85$ ($\pm10$)\\
$0.95$ & $12090$ ($\pm5894$) & $80$ ($\pm10$) & $7617$ ($\pm5696$) & $83$ ($\pm10$)\\
\hline
\end{tabular}
}
\end{scriptsize}
\end{center}
\end{table*}

The results in Table~\ref{lscale} show that without a dual memory system, $\alpha$ must be extremely high ($\alpha=0.95$) for the model to successfully learn Boxing to a similar standard as RePR. However, both models show that high $\alpha$ values result in considerable forgetting of Road Runner. Importantly, this means that without a dual memory system there does not exist an $\alpha$ value that results in both successful learning of the new task and acceptable retention of the previous task. When using RePR's dual memory system, learning the new task is considerably easier and thus, comparably lower $\alpha$ values will result in both successfully learning the new task and retaining the previous task. Overall, these results suggest that omitting the dual memory system dramatically increases the interference between new knowledge and knowledge of previous tasks.

\begin{algorithm}[p]
\label{stm-alg}
\DontPrintSemicolon
\SetKwProg{Def}{def}{:}{end}

\Def{\upshape train\_stm($env$, $exp$)}{
  $stm\_agent$ = init\_stm\_agent()\;
  $exp$.clear()\;
  \While{\upshape training $stm\_agent$}{
    $exp$.add($env$.step($stm\_agent$.pred\_net.calc\_action()))\;
    $a_t$, $r_t$, $d_t$, $s_t$, $s_{t+1}$ = $exp$.sample()\;
    \eIf{\upshape $d_t$}{
      $y_t$ = $r_t$\;
    }{
      $y_t$ = $r_t + {\gamma}\max\limits_{a_{t+1}} stm\_agent\text{.target\_net.get\_outputs(}s_{t+1}\text{)}[a_{t+1}]$\;
    }
    $loss$ = $(y_t - stm\_agent\text{.pred\_net.get\_outputs(}s_t\text{)}[a_t])^2$\;
    $stm\_agent$.pred\_net.SGD\_step($loss$)\;
    \If{\upshape $iter$ \textbf{divisible by} $update\_target\_freq$}{
      $stm\_agent$.target\_net = copy($stm\_agent$.pred\_net)\;
    }
  }
  \KwRet $stm\_agent$\;
}
\caption{Simplified pseudo-code demonstrating the training procedure of the short-term agent in RePR. This pseudo-code implements the loss function defined in Equation~\ref{dqn-loss}. The $stm\_agent$ contains a predictor network and a target network; $env$ is the environment currently being learnt; $exp$ is an experience replay; $iter$ is the current training iteration; $update\_target\_freq$ is the frequency at which the target network is updated. In practice, the loss function is calculated by averaging over a batch of samples.}
\end{algorithm}

\begin{algorithm}[p]
\label{ltm-alg}
\DontPrintSemicolon
\SetKwProg{Def}{def}{:}{end}
\Def{\upshape train\_ltm($env$, $exp$, $stm\_agent$, $ltm\_agent$, $gan$)}{
  $exp$.clear()\;
  $prev\_ltm\_agent$ = copy($ltm\_agent$)\;
  \While{\upshape training $ltm\_agent$}{
    $exp$.add($env$.step($ltm\_agent$.pred\_net.calc\_action()))\;
    $a_t$, $r_t$, $d_t$, $s_t$, $s_{t+1}$ = $exp$.sample()\;
    $z$ = uniform\_sample($-1$, $1$, $n\_latents$)\;
    $\widetilde{s}$ = $gan$.gen.get\_outputs($z$)\;
    $L_D$ = $\text{sum(}(ltm\_agent\text{.pred\_net.get\_outputs(}s\text{)} - stm\_agent\text{.pred\_net.get\_outputs(}s\text{)})^2\text{)}$\;
    $L_{PR}$ = $\text{sum(}(ltm\_agent\text{.pred\_net.get\_outputs(}\widetilde{s}\text{)} - prev\_ltm\_agent\text{.pred\_net.get\_outputs(}\widetilde{s}\text{)})^2\text{)}$\;
    $loss$ = $\alpha L_D + (1 - \alpha) L_{PR}$\;
    $ltm\_agent$.pred\_net.SGD\_step($loss$)\;
  }
  \KwRet $ltm\_agent$\;
}
\caption{Simplified pseudo-code demonstrating the training procedure of the long-term agent in RePR. This pseudo-code implements the loss function defined in Equation~\ref{repr-loss}. The $stm\_agent$ contains a predictor network; the $ltm\_agent$ contains a predictor network; the $gan$ contains a generator network; $env$ is the environment currently being learnt; $exp$ is an experience replay; $\alpha$ weights the importance of learning the new task vs. retaining previously learnt tasks; $n\_latents$ is the number of latent input variables used by the GAN. In practice, the loss function is calculated by averaging over a batch of samples and pseudo-items are sampled from a large array of items generated by the GAN before training the agent.}
\end{algorithm}

\begin{algorithm}[p]
\label{gan-alg}
\DontPrintSemicolon
\SetKwProg{Def}{def}{:}{end}
\Def{\upshape train\_gan($exp$, $gan$)}{
  $new\_gan$ = initialise\_gan()\;
  \While{\upshape training $new\_gan$}{
    \eIf{\upshape $iter$ \textbf{is} even}{
      \eIf{\upshape True with probability of $1/T$}{
        $a_t$, $r_t$, $d_t$, $s_t$, $s_{t+1}$ = $exp$.sample()\;
        $x$ = $s_t$
      }{
        $z$ = uniform\_sample($-1$, $1$, $n\_latents$)\;
        $\widetilde{s}$ = $gan$.gen.get\_outputs($z$)\;
        $x$ = $\widetilde{s}$
      }
      $z$ = uniform\_sample($-1$, $1$, $n\_latents$)\;
      $\epsilon$ = uniform\_sample($0$, $1$)\;
      $\widetilde{x}$ = $new\_gan$.gen.get\_outputs($z$)\;
      $\hat{x}$ = $\epsilon x + (1 - \epsilon)\widetilde{x}$\;
      $disc\_real$ = $new\_gan$.disc.get\_outputs($x$)\;
      $disc\_fake$ = $new\_gan$.disc.get\_outputs($\widetilde{x}$)\;
      $disc\_xhat$ = $new\_gan$.disc.get\_outputs($\hat{x}$)\;
      $gradient\_penalty$ = $\lambda(\norm{\text{grads(}disc\_xhat, \hat{x}\text{)}}_2-1)^2$\;
      $loss$ = $disc\_fake - disc\_real + gradient\_penalty + \epsilon_{drift} disc\_real^2 + \epsilon_{drift} disc\_fake^2$\;
      $new\_gan$.disc.SGD\_step($loss$)\;
    }{
      $z$ = uniform\_sample($-1$, $1$, $n\_latents$)\;
      $\widetilde{x}$ = $new\_gan$.gen.get\_outputs($z$)\;
      $loss$ =  $-new\_gan$.disc.get\_outputs($\widetilde{x}$)\;
      $new\_gan$.gen.SGD\_step($loss$)\;
    }
  }
  \KwRet $new\_gan$\;
}
\caption{Simplified pseudo-code demonstrating the training procedure of the long-term GAN in RePR. This pseudo-code implements the training data selection procedure defined in Equation~\ref{gan-data} and the loss functions defined in Equation~\ref{disc-loss} and Equation~\ref{gen-loss}. The $gan$ contains a generator network; the $new\_gan$ contains both a discriminator network and a generator network; $exp$ is an experience replay containing samples from the current environment; $iter$ is the current training iteration; $T$ is the number of tasks seen; $n\_latents$ is the number of latent input variables used by the GAN. In practice, the loss functions are calculated by averaging over a batch of samples and pseudo-items are sampled from a large array of items generated by the previous GAN before training the new GAN.}
\end{algorithm}

\section*{Acknowledgment}
We gratefully acknowledge the support of NVIDIA Corporation with the donation of the TITAN X GPU used for this research. We also wish to acknowledge the use of New Zealand eScience Infrastructure (NeSI) high performance computing facilities. New Zealand's national facilities are provided by NeSI and funded jointly by NeSI's collaborator institutions and through the Ministry of Business, Innovation \& Employment's Research Infrastructure programme. URL \url{https://www.nesi.org.nz}.

\section*{Declaration of Competing Interest}
The authors declare that they have no known competing financial interests or personal relationships that could have appeared to influence the work reported in this paper.

\newpage

\bibliographystyle{templates/elsarticle-num} 
\bibliography{mybibfile}






\phantom\newline\phantom\newline

\begin{wrapfigure}{l}{25mm}
	\includegraphics[width=1in,height=1.25in,clip,keepaspectratio]{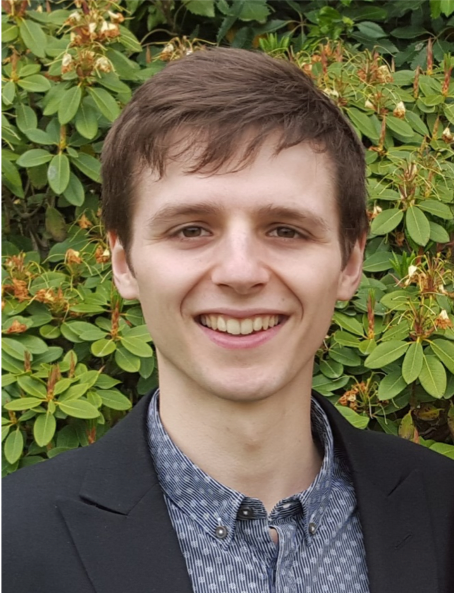}
\end{wrapfigure}\par
\textbf{Craig Atkinson} received his B.Sc. (Hons.) from the University of Otago, Dunedin, New Zealand, in 2017. He has just completed his doctorate in Computer Science at the University of Otago. His research interests include deep reinforcement learning and continual learning.
\newline\newline\newline

\begin{wrapfigure}{l}{25mm}
	\includegraphics[width=1in,height=1.25in,clip,keepaspectratio]{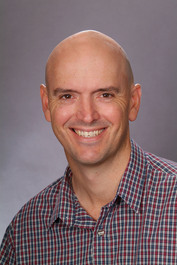}
\end{wrapfigure}\par
\textbf{Brendan McCane} received the B.Sc. (Hons.) and Ph.D. degrees from the James Cook University of North Queensland, Townsville City, Australia, in 1991 and 1996, respectively. He joined the Computer Science Department, University of Otago, Otago, New Zealand, in 1997. He served as the Head of the Department from 2007 to 2012. His current research interests include computer vision, pattern recognition, machine learning, and medical and biological imaging. He also enjoys reading, swimming, fishing and long walks on the beach with his dogs.
\newline

\begin{wrapfigure}{l}{25mm}
	\includegraphics[width=1in,height=1.25in,clip,keepaspectratio]{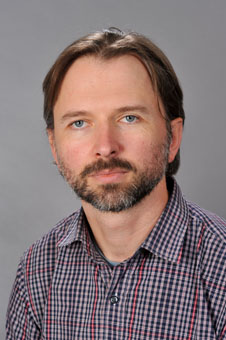}
\end{wrapfigure}\par
\textbf{Lech Szymanski} received the B.A.Sc. (Hons.) degree in computer engineering and the M.A.Sc. degree in electrical engineering from the University of Ottawa, Ottawa, ON, Canada, in 2001 and 2005, respectively, and the Ph.D. degree in computer science from the University of Otago, Otago, New Zealand, in 2012.   He is currently a Lecturer at the Computer Science Department at the University of Otago. His research interests include machine learning, artificial neural networks, and deep architectures.
\newline

\begin{wrapfigure}{l}{25mm}
	\includegraphics[width=1in,height=1.25in,clip,keepaspectratio]{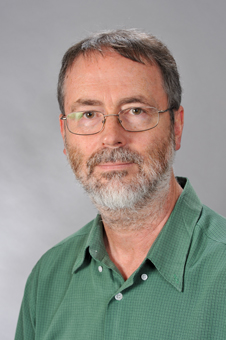}
\end{wrapfigure}\par
\textbf{Anthony Robins} completed his doctorate in cognitive science at the University of Sussex (UK) in 1989.  He is currently a Professor of Computer Science at the University of Otago, New Zealand.  His research interests include artificial neural networks, computational models of memory, and computer science education.

\end{document}